\documentclass[letterpaper, 10 pt, journal, twoside]{IEEEtran}
\usepackage{amsmath,amssymb,amsfonts}
\usepackage{graphicx}
\usepackage{xcolor}
\usepackage{hyperref} 
\hypersetup{
	colorlinks=true,
	linkcolor=[rgb]{0,0.4,0.4},
	filecolor=black,      
	citecolor=blue,
	urlcolor=cyan,
}

\usepackage{textcomp}
\usepackage[center]{subfigure}
\usepackage{multirow}
\usepackage{bm}
\usepackage{siunitx}
\usepackage{amsthm}
\usepackage{nicefrac} 

\usepackage{adjustbox}

\renewcommand{\v}[1]{{\boldsymbol{\mathbf{#1}}}}
\newcommand\state{\v{x}}
\newcommand\latent{\v{\Tilde{x}}}
\newcommand\desiredlatent{\v{\Tilde{x}_d}}
\newcommand\obs{\v{y}}
\newcommand\causes{\v{v}}

\newcommand\action{\v{u}}
\newcommand\g{g}
\newcommand\f{f}
\newcommand\ynoise{\v{z}}
\newcommand\munoise{\v{w}}

\newcommand\KL{\text{\bf{KL}}}

\usepackage{rotating}
\usepackage{booktabs}
\usepackage{algorithm}
\usepackage[noend]{algpseudocode}

\usepackage{array}
\newcolumntype{L}[1]{>{\raggedright\let\newline\\\arraybackslash\hspace{0pt}}m{#1}}
\newcolumntype{C}[1]{>{\centering\let\newline\\\arraybackslash\hspace{0pt}}m{#1}}
\newcolumntype{F}[1]{>{\let\newline\\\arraybackslash\hspace{0pt}}m{#1}}

\graphicspath{ {./img/} }
\begin{document}
%
\title{Active Inference in Robotics and Artificial Agents: Survey and Challenges}
%
%
%

\author{Pablo~Lanillos,
        Cristian~Meo,
        Corrado~Pezzato,
        Ajith~Anil~Meera,
        Mohamed~Baioumy,
        Wataru~Ohata,
        Alexander~Tschantz,
        Beren~Millidge,
        Martijn~Wisse,
        Christopher~L.~Buckley,
        and~Jun~Tani

\thanks{P. Lanillos is with the Donders Insitute for Brain Cognition and Behavior, Dept. of Artificial Intelligence, Radboud University. Netherlands.}%
\thanks{C. Meo, C. Pezzato, A. A. Meera and M. Wisse are with the Robotics Institute, Delft University of Technology. Netherlands.}%
\thanks{B. Millidge is with the MRC Brain Networks Dynamics Unit, Oxford University, UK.}%
\thanks{M. Baioumy is with the Oxford Robotics Institute, Oxford University, UK}
\thanks{A. Tschantz and C. L. Buckley are with the Dept. of Informatics, University of Sussex, UK.}%
\thanks{W. Ohata and J. Tani are with with the Okinawa Institute of Science and Technology, Japan.}
\thanks{This manuscript is currently under review.}}

%
%

\markboth{IEEE Template. Submitted version}%
{Lanillos \MakeLowercase{\textit{et al.}}: Active Inference in Robotics and Artificial Agents: Survey and Challenges}
%


\maketitle

\begin{abstract}
Active inference is a mathematical framework which originated in computational neuroscience as a theory of how the brain implements action, perception and learning. Recently, it has been shown to be a promising approach to the problems of state-estimation and control under uncertainty, as well as a foundation for the construction of goal-driven behaviours in robotics and artificial agents in general. Here, we review the state-of-the-art theory and implementations of active inference for state-estimation, control, planning and learning; describing current achievements with a particular focus on robotics. We showcase relevant experiments that illustrate its potential in terms of adaptation, generalization and robustness. Furthermore, we connect this approach with other frameworks and discuss its expected benefits and challenges: a unified framework with functional biological plausibility using variational Bayesian inference.
\end{abstract}


\IEEEpeerreviewmaketitle



\section{Introduction}
\label{sec:intro}

\IEEEPARstart{T}{his} survey presents the current work in active inference for robotics and artificial agents and discusses the benefits and challenges it must address if it is to become a practical and revolutionary unified mathematical framework for estimation, control, planning and learning.

\begin{figure}[t!]
	\centering
    \includegraphics[width=0.95\linewidth]{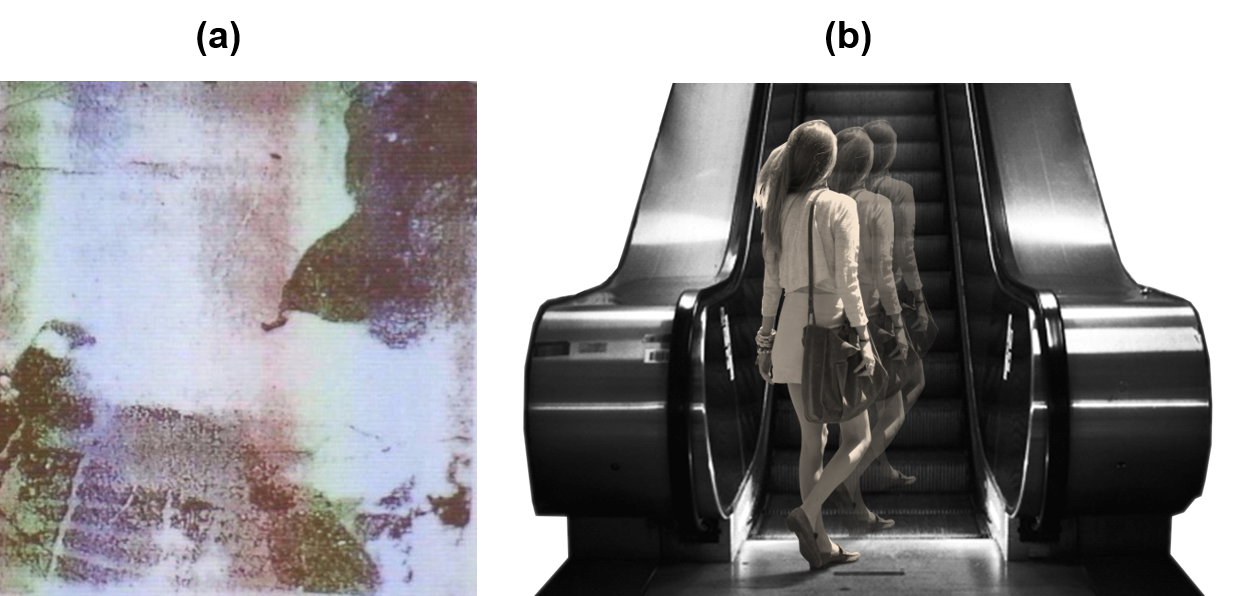}
	\caption{\textbf{Predictive brain approach}. (a) Dallenbach illusion~\cite{kmd1951puzzle} and the importance of prior information in perception. (b) The escalator effect. Involuntary active strategies triggered by a predicted world state using prior knowledge acquired through experience.}
	\label{fig:intro_motivation}
	\vspace{-10px}
\end{figure}

Active inference (AIF) is a biologically plausible mathematical construct based on the free energy principle (FEP) proposed by Karl Friston~\cite{friston2010free}. This principle describes how living systems resist a natural tendency to disorder. Its backbone can be traced back to the work of Helmholtz on perception~\cite{Helmholtz1867}. In the presence of uncertain stimuli, such as the Dallenbach illusion~\cite{kmd1951puzzle} depicted in Fig.~\ref{fig:intro_motivation}a, our perceptual apparatus tends to fill in for missing information by utilizing prior knowledge, a process referred to as  \textit{unconscious inference}~\cite{Helmholtz1867}. Fig.~\ref{fig:intro_motivation}a shows a cow rotated 90 degrees clockwise that is impossible to see before the presence of this prior information and impossible to un-see afterwards. One implication of this theory is that it suggests the brain maintains an internal model, i.e., a generative model, of the causes of sensation which is combined with the sensory stream to form a given percept. Algorithmically it has been suggested this is achieved by minimizing the difference (error) between top-down predictions of this internal model and incoming sensory data. Importantly, under this predictive coding framework~\cite{rao1999predictive}, it has also been suggested that agents can also act on the world to change the sensation to better fit the same internal model thus providing a dual account of both perception and action~\cite{buckley2017free,lanillos2021neuroscience}. Again phenomenal experience is supportive of this active component~\cite{lanillos2020predictive}. For example when reaching an escalator (Fig.~\ref{fig:intro_motivation}b), even if it is broken, we perceive it moving and we prepare our body to fit the velocity of the stair and are often surprised. Once we realize that it is stopped, we adapt again to the new situation.

Hence, adaptive behaviour can be viewed as an active inference process in which the agent selects those actions that support the maximization of model evidence, or equivalently, the minimization of surprise. This can be performed by exploiting the internal model or generating exploratory behaviours that reduce the model entropy: i.e., reduce uncertainty by maximizing information gain. This work describes how these concepts can be applied to and enrich robotic systems.

\subsection{Overview}
\label{sec:sota_list}

We formalize and describe AIF in the context of current challenges in robotics, where adaptation to uncertain, complex and changing environment plays a major role. We survey robotics applications of AIF and provide the appropriate mathematical and theoretical background. Hence, aiming to offer both a review and a technical reference. Table \ref{tab:overview} summarises the most relevant works organized into: state-estimation, control, planning (i.e., computing actions into the future), and high-level cognitive skills (e.g., self-other distinction).

\begin{table*}[hbtp!]
\caption{Overview Active Inference (AIF) state-of-the-art in robotics and control}
\centering 
\begin{tabular}{C{0.5cm}| L{4.5cm}| L{7.0cm} | L{3cm}} %
\toprule
\midrule
\multicolumn{2}{c|}{\textbf{Research Topic}}  & \textbf{Approach and Implementation} & \textbf{References}\\ [0.5ex] %
\hline
\vspace{5mm}
\multirow{9}{*}{\begin{turn}{90}\textbf{Estimation}\end{turn}} &  Linear systems with colored noise & Dynamic Expectation Maximization & \cite{e23101306}, \cite{meera2020free}, \cite{ajith2021DEM_drone}, \cite{bos2021free} \\
\cline{2-4} \vspace{5mm} 
& Multi-sensory estimation and learning & Predictive coding & \cite{lanillos2018adaptive,oliver2021empirical}\\ \cline{2-4}  \vspace{5mm}
& Localization & Laser-based continuous AIF & \cite{burghardt2021robot}  \\ 

\hline 
\hline
\vspace{4mm}
\multirow{8}{*}{\begin{turn}{90}\textbf{Control} \end{turn}} 
& \multirow{3}{*}{Humanoid robots and manipulators} &  Continuous torque control. Low dimensional input & \cite{pio2016active}, \cite{Lanillos2018ActiveIW}, 
\cite{pezzato2020novel}, \cite{baioumy2020active}\\ \cline{3-4} \vspace{4mm}

& & High-dimensional input with function learning (Deep AIF) & \cite{sancaktar2020end}, \cite{meo2021multimodal}, \cite{rood2020deep}\\ \cline{2-4} \vspace{4mm}

&  Fault-tolerant systems & Threshold on the surprise, precision learning  &  \cite{baioumy2021fault}, \cite{pezzato2020active}  \\
\cline{2-4} \vspace{4mm}

& Bio-inspired agents & Phototaxis, Continuous AIF  & \cite{Baltieri2017Active} \\ 

\hline
\hline
\multirow{10}{*}{\begin{turn}{90}\textbf{Planning}\end{turn}} 

& \multirow{8}{*}{Discrete-time stochastic control} \vspace{6mm} & AIF with rewards &  \cite{friston2009reinforcement}, \cite{millidge2020reinforcement}, \cite{sajid2019active}, \cite{tschantz2020scaling}, \cite{Han-Tani-RL-AIF-2021},
 \cite{millidge2019combining}\\ \cline{3-4} \vspace{3mm}
& & Discrete Inference & \cite{tschantz2020learning}, \cite{friston2020sophisticated} \\ \cline{3-4} \vspace{3mm}
& & Deep AIF & \cite{ueltzhoffer2018deep}, \cite{van_der_Himst_2020},   \cite{fountas2020deep}, \cite{millidge2020deep} \\ \cline{2-4} \vspace{3mm}
&  \multirow{3}{*}{Navigation} & Deep AIF &  \cite{ccatal2020learning}, \cite{catal2020deep}, \cite{catal2021robot}, \cite{Matsumoto_2020} \\ \cline{3-4} \vspace{3mm}

& & Recurrent spiking neural networks & \cite{Traub2021Dynamic} \\
\hline 
\hline
\vspace{6mm}
\multirow{7}{*}{\begin{turn}{90}\textbf{Cognitive}\end{turn}}

& Symbolic reasoning & AIF + behavior trees &  \cite{pezzato2020activeBT} \\ \cline{2-4} \vspace{5mm}

& Human robot interaction & Imitative interactions via visio-proprioceptive sequences & \cite{murata-tani2015}, \cite{Ohata_2020}, \cite{chame2020AHybrid}, \cite{chame2020cognitive}  \\
\cline{2-4} \vspace{4mm}
& Self/other distinction - Self-recognition & {Robot mirror test using movement and visual cues} & \cite{lanillosRobotSelfOther2020, hoffmann2021robot} \\
\hline 
\hline 
\end{tabular}
\label{tab:overview}
\end{table*}

\subsection{Paper structure}
\label{sec:structure}

Section \ref{sec:aif} introduces the common ground and notation. State-estimation, control, planning, learning, and hierarchical representations are mathematically described in Sec. \ref{sec:aif_estimation}, Sec. \ref{sec:aif_control} and Sec. \ref{sec:aif_planning} respectively. Section \ref{sec:applications} details relevant robotic experiments that showcase the advantages of AIF approaches. Finally, Sec. \ref{sec:connect} describes the relation of AIF with other frameworks, such as classical control and reinforcement learning, and Sec. \ref{sec:conclusion} discusses the benefits and challenges to make AIF a standard modelling technique for robotic systems.

\begin{figure*}[hbtp!]
	\centering
    \includegraphics[width=0.95\linewidth]{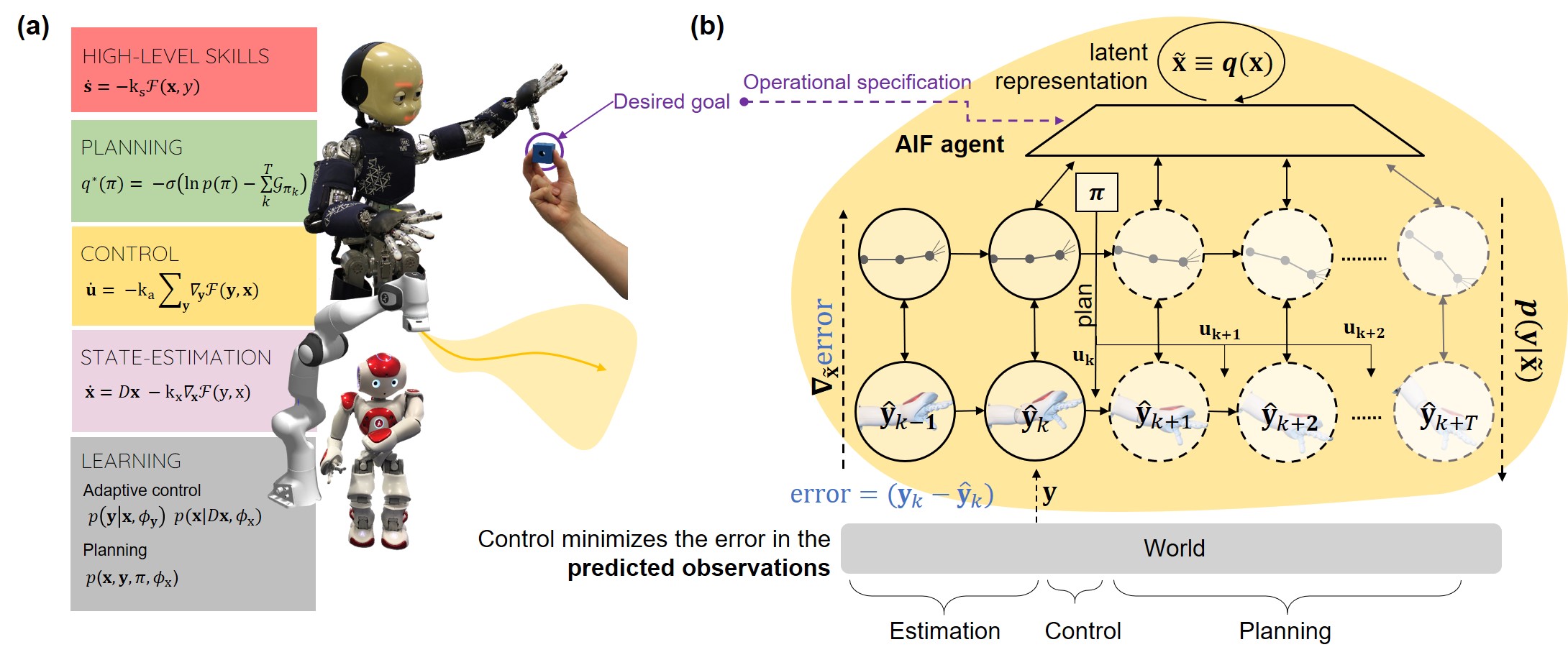}
	\caption{\textbf{Active inference in robotics}. (a) Estimation, control, planning and learning using the AIF framework. (b) Hierarchical abstract description, error message passing and subdivision of estimation (using past information), adaptive control and planning of the future predictions and actions.}
	\label{fig:intro_structure}
\end{figure*}

\section{Active inference common ground}
\label{sec:aif}
We introduce the standard equations and concepts from the AIF literature, and the notation used in this paper, framed for estimation and control of robotic systems. 

AIF solves the dual problem of estimation and control by optimizing a single objective: a free energy bound~\cite{friston2010free,buckley2017free,lanillos2021neuroscience}. This entails updating the internal state and generating the control actions that minimize the error in the predicted observations. Hence, AIF has the particularity that the generative model of actions is cast into a generative model of predicted observations or inputs. In contrast to other similar approaches, control actions are generated to make the world more predictable and the least surprising. Figure~\ref{fig:intro_structure}b sketches the hierarchical AIF, where the lowest level is the sensory input. Estimation and control is solved through Bayesian inference, e.g., usually using stochastic variational approaches and transforming the inference problem into an optimization problem. In AIF the internal state is conditionally independent of the external state (world). However, they can affect each other through sensory and action states~\cite{kirchhoff2018markov}. An important consequence is that the robot encodes as preferences its intentions and these preferences drive both the state-estimation and the control.

To help new robotic researchers in AIF to get started, we provide a very distilled summary of works in Table \ref{tab:getstarted}. It contains what the authors consider seminal papers that led to the current advancements in the state of the art with the focus on the control and robotics communities.

\begin{table}[ht]
\caption{Getting started with Active Inference}
\centering 
\begin{tabular}{L{5.0cm} L{2cm}} %
\toprule
\midrule
\textbf{Topic} & \textbf{References}\\ [0.5ex] %
\midrule 
General introduction and tutorial & \cite{bogacz2017tutorial,buckley2017free}, \\
Derivations for robot control & \cite{oliver2021empirical, pezzato2020novel} \\
Relationship with classical control & \cite{baltieri2019pid, baioumy2020active} \\
Relationship with optimal control & \cite{vandelaar2020application,2020ICRA_baioumy} \\
Discrete Active Inference & \cite{da2020active}\\
RL and active inference & \cite{tschantz2020reinforcement}, \cite{friston2009reinforcement} \\
State estimation & \cite{Friston2008DEM}, \cite{meera2020free}\\
Predictive processing & \cite{ciria2021predictive}, \cite{Spratling2017review} \\
Human Robot Interaction & \cite{chame2020AHybrid} \\
Neuroscientific foundations & \cite{friston2010free}, \cite{PEZZULO201517} \\
\hline 
\end{tabular}
\label{tab:getstarted}
\end{table}

For consistency, we will use notation described in Table~\ref{tab:notation}. Figure \ref{fig:intro_structure}b describes the AIF architecture with the sensory observations $\obs$ and the latent states $\latent$. Variables can be matrices or tensors. For instance, the sensory observation $\obs$ may have the dimension of every sensor modality (e.g. visual and joints angles) and their higher-order derivatives\footnote{In the AIF terminology this way to encode observation and states is called generalized coordinates---See Appendix~\ref{appendix:gen-coords} for a detailed explanation.}---for rotational joints these are the angles, the joints velocities, accelerations, etc.

\begin{table}[hbtp!]
\caption{Notation}
\label{tab:notation}
\begin{adjustbox}{width=\columnwidth,center}
\begin{tabular}{cc|cc} %
\toprule
\midrule
\textbf{Var.} & \textbf{Definition} & \textbf{Function[al]} & \textbf{Definition}\\ [0.5ex] %
\midrule 
 $\state$ & Process state &  $\g$ & Observation model\\
 $\latent$ & Internal/inferred state & $\f$ & Dynamic/transition model\\
 $\obs$ & Observations &  $\Pi=\Sigma^{-1}$ & Precision/Inverse variance\\
 $\causes$ & General causes & $\nabla_{\Tilde{\state}}$ &  Partial derivative w.r.t. state \\
 $\action$ & Control actions & $q(\state)$ &  Variational density of the state \\
 $\ynoise$ & Observation noise & $\mathcal{F}$ & Variational free energy\\
 $\munoise$ & Internal state noise & $\KL$ & Kullback–Leibler divergence\\
\hline 
\end{tabular}%
\end{adjustbox}
\vspace{-10px}
\end{table}

\subsection{From Bayesian Inference to the Free Energy Principle}
We will start by designing an agent that does not have access to the world/body state but has to infer it from the sensor measurements. In terms of Bayesian inference, it infers the most probable state of the world $\state$ using imperfect noisy sensory observations $\obs$. According to Bayes rule,

\begin{equation}
    p(\state|\obs) = \frac{p(\state, \obs)}{p(\obs)} =\frac{p(\obs|\state)p(\state)}{p(\obs)}.
    \label{eq:bayesrule}
\end{equation}

The probability of a state $\state$ given the observed data $\obs$ is encoded in the \textit{posterior probability} $p(\state|\obs)$. The \textit{likelihood} $p(\obs|\state)$ measures the compatibility of the sensory input with the state, while the \textit{prior probability} $p(\state)$ is the current belief about the state before receiving the observation $\obs$. Finally, $p(\obs)$ is the \textit{marginal likelihood}, which corresponds to the probability of observing $\obs$ regardless of the state. The goal is to find the value of $\state$ which maximizes the posterior.

\subsubsection{Variational Free Energy}
Now, we model the influence of the world on the agent as the tendency of the agent to find an equilibrium between its internal model and the external process. This is the core of the free energy principle \cite{friston2010free}, which in Bayesian terminology states that the agent maximizes model evidence by approximating the internal state described by the density $q(\state)$ to the world posterior $p(\state|\obs)$. Besides the biological motivation, there is also a computational reason  for introducing this variational density. The posterior $p(\state|\obs)$ is typically intractable and cannot be evaluated directly in most cases, particularly in continuous spaces. In the free energy principle, a variational Bayes approach is used to obtain a tractable solution\footnote{The variational inference approach is common in modern machine learning for approximating probability densities~\cite{jordan1999introduction}.}. Instead of computing an exact posterior, it is approximated through optimization~\cite{Friston2007Variational,buckleyFEP}. This approach requires the auxiliary variational density $q(\state)$. The idea is to minimize the \textit{Kullback-Leibler divergence} $(\KL)$ between $q(\state)$ and $p(\state|\obs)$, as the divergence will approach to 0 when both distributions are the same.
\begin{align}
\label{eq:D_KL}
\KL\left[ q(\state) || p(\state|\obs)) \right] &= \int q(\state) \log \frac{q(\state)}{p(\state|\obs)}d\state \nonumber\\
&=-\int q(\state) \log \frac{q(\state)}{p(\obs, \state)}d\state+\log{p(\obs)}  \nonumber\\
&= \mathcal{F} + \log p(\obs) \geq 0 .
\end{align}
$\mathcal{F}$ is defined as the variational free energy\footnote{In machine learning the negative VFE is also known as the Evidence Lower Bound (ELBO). See Appendix~\ref{appendix:vfe} for demonstration.} (VFE) and measures the divergence between the variational density $q(\state)$ and the joint distribution (generative model) $p(\obs,\state)$. The VFE can be evaluated because it depends on $q(\state)$ and the knowledge about the environment of the agent $p(\obs, \state) = p(\obs|\state)p(\state)$. 

Interestingly, because the $\KL$ is always positive the VFE is an upper bound on the surprise: $\mathcal{F} \geq - \log p(\obs)$, which measures the atypicality of events quantified through the negative log probability of sensory data. Therefore, optimizing $\mathcal{F}$ is equivalent to evaluating the posterior density. In the ideal case (e.g., no noise), when the model is able to capture the real generative process $\KL\left[ q(\state) || p(\state|\obs)) \right]$ is zero and $\mathcal{F}$ becomes the marginal likelihood or surprise. The key advantage of this formalism is that it reduces the intractable Bayesian inference problem given in Eq.~\ref{eq:bayesrule} into an optimization problem. \textit{Crucially for AIF agents, the state and action are simultaneously inferred by optimizing $\mathcal{F}$}.

\subsubsection{Mean-field and Laplace approximation}
We did not describe yet the nature of the auxiliary density $q(\state)$ that is encoded by the internal state of the system. Here, for mathematical convenience and to reflect the majority of the active inference literature, we choose $q(\state)$ as a factorization of random variables with known form, i.e., Gaussian $q(\state) \sim \mathcal{N}(\state|\latent, \Sigma_Q )$, and track the densities with the sufficient statistics defined by the mean and the variance (see Appendix \ref{appendix:laplace} and \ref{appendix:mean-field}). Other forms of variational approximations are out of the scope of this paper. The VFE ($\mathcal{F}$ in Eq. \ref{eq:D_KL}) is by definition:
\begin{align}
    \mathcal{F} &= \int q(\state) \log \frac{q(\state)}{p(\obs, \state)}d\state\\
    &=\int q(\state) \log q(\state)d\state-\int q(\state)p(\obs, \state)d\state.
    \label{eq:vfe}
\end{align}
The VFE under the mean-field and Laplace approximations simplifies to\footnote{Note that the first term of Eq. (\ref{eq:vfe}) vanishes and the second term becomes Eq. \ref{eq:free_energy}. See Appendix~\ref{appendix:mean-field} for a short explanation of the mean field and Laplace approximations.}:
\begin{align}
    \mathcal{F} =- \log p(\obs, \latent) -\frac{1}{2}\log{(2\pi\Sigma_Q)}.
    \label{eq:free_energy}
\end{align}
where $\latent$ are the sufficient statistics of the factorized variational density that codifies the process state $\state$, and $\Sigma_Q$ is the optimal variance that optimizes the variational free energy.

One of the advantages of using these approximations is that estimation and control becomes a quadratic optimization problem that explicitly minimizes the model error prediction. We will show this explicitly in the following sections.

\subsection{Generative models}
\label{sec:aif_state}
We differentiate two functionals~\cite{friston2010action}: the generative process, which defines the real system in the environment that is responsible for data generation---usually referred to as 'the plant' in control engineering--- and the generative model, which describes the agent's internal representation (approximation) of the generative process. The AIF agent's behaviour is driven by the generative model. This model can be defined by an expert designer (operational specification) or learnt through interaction with the world.

The generative model of the system at instant $k$ given all past observations and states is defined as the joint distribution over states and observations:
\begin{align}
p(\obs_{0:T}, \latent_{0:T}) = \prod_{k=0}^T \textcolor{blue}{p(\obs_k | \latent_k)} \textcolor{red}{ p(\latent_k | \latent_{k-1})}
\label{eq:gen_model1}
\end{align}
Where the transition model $p(\latent_k | \latent_{k-1})$ collapses to $p(\latent_0)$ when $k=0$,  $p(\obs_k | \latent_k)$ is the likelihood of the observations given the state (observation model).

We can also describe the generative model from the dynamical systems approach using state-space equations:
\begin{subequations}
\begin{align}
    \dot{\Tilde{\state}} = D\Tilde{\state} &= \f(\Tilde{\state},\causes) + \munoise & \text{internal state dynamics} \label{eq:statespace:dynamics}\\
        \obs &= \g(\Tilde{\state}) + \ynoise & \text{observation model} \label{eq:statespace:sensory}
\end{align}
\end{subequations}
Both equations encode the generative model described in Eq.~\ref{eq:gen_model1}. Equation~\ref{eq:statespace:dynamics} describes the evolution of the internal state---using the prior information or desired preference---and Eq.~\ref{eq:statespace:sensory} describes the generative model of the causes, usually simplified to the likelihood of the sensory output given the internal states. Here $\Tilde{\state}$, $\obs$ are the estimated state and output, and $\munoise$ and $\ynoise$ are the process and observation (Gaussian) noise. Finally, $D\latent=\frac{d\latent}{dt}$ is the time-derivative of the state vector\footnote{An explanation of this operation and the generalized coordinates used in continuous AIF can be found in the Appendix~\ref{appendix:gen-coords}.}.


The functions that describe the generative model $g(\Tilde{\state})$ and $f(\Tilde{\state}, \causes)$ can be explicit or function approximators, such as neural networks. According to the model type chosen, different behaviours of an agent can be achieved.

As a useful and particular example, in the case of a linear plant, the agent's generative model can be described as:
\begin{align}
    \dot{\Tilde{\state}} &= A\Tilde{\state} + B\causes + \munoise;\quad\quad
        \obs = C\Tilde{\state} + \ynoise, \nonumber \label{eq:statespace:linear}
\end{align}
where $A$, $B$ and $C$ are the plant specific matrices and $\causes$ here is the control input.

\subsection{Perception and action}
As stated at the beginning of Sec.~\ref{sec:aif}, AIF agents perceive and act in the environment by optimizing the same objective, i.e., the VFE~\cite{friston2010free,oliver2021empirical}. Perception involves the estimation of states $\latent$ and parameters (e.g., $A,B,C$), whereas action has a dual role of resolving uncertainty and obtaining new data which concords with the agent’s belief/intention. Action involves the computation of the control signal $\action$ to act in the environment. Together, the optimization of VFE through perception and action drives the agent towards reducing the prediction error and producing better sensory predictions. Assuming that the variational density is described by the latent variable $\latent$ they solve the following equations:
\begin{subequations}
\begin{align}
\latent &= \arg\min_\latent \mathcal{F}(\latent,\obs, \causes) 	\\
\action &= \arg\min_\action \mathcal{F}(\latent, \obs(\action),\causes)
\end{align}
\end{subequations}
Both equations are usually solved through gradient descent. It is important to highlight that original works on active inference do not explicitly model the action in the generative model as it is encoded within the observation $\obs(\action)$. Alternatives on this formulation have been left for the discussion section.

\section{State-estimation}
\label{sec:aif_estimation}
This section summarizes the mathematical formalization of the first of the four blocks treated in this paper: state-estimation. We solve state-estimation similarly to applying a Bayesian filter. This involves the estimation of two components: the mean estimate and the associated confidence (precision) of the estimate. Under the free energy principle framework, both can be estimated using the first two gradients of the free energy. State inference is driven by the following differential equation:
\begin{equation}
   \boldsymbol{\dot{\Tilde{\state}}} =  D\Tilde{\state} - \kappa_{x}\nabla_{\Tilde{\state}}\mathcal{F}(\latent,\obs)
    \label{eqn:state_update_rule}
\end{equation}
The VFE under the Laplace and mean-field approximations has closed form and is defined as:
\begin{align}
\nonumber
    \mathcal{F}(\latent,\obs) \triangleq & -\ln p(\latent,\obs) = -\ln p(\obs|\latent) p(\latent)\\
    \triangleq& \;(\obs-g(\latent))^T \Sigma_{\ynoise}^{-1} (\obs-g(\latent)) \nonumber\\
    &+ (D\latent-f(\latent,\causes))^T \Sigma_{\munoise}^{-1}(D\latent-f(\latent,\causes)) \nonumber\\
    &+ \frac{1}{2} \ln|\Sigma_{\munoise}| + \frac{1}{2} \ln|\Sigma_{\ynoise}| \label{eq:flaplace}
\end{align}
\subsection{State-estimation minimizes the prediction error}

By defining $\bm \varepsilon_{\obs}=(\obs-g(\latent))$ and $\bm \varepsilon_\latent=(D\latent-f(\latent,\causes))$ as the sensory and state model prediction errors in Eq.~\ref{eq:flaplace}, $\nabla_{\Tilde{\state}}\mathcal{F}$ becomes the gradient of the weighted sum of squared prediction errors. This is very relevant as it connects AIF with the theory of predictive coding in the brain~\cite{friston2009predictive}.
\begin{align}
\label{eq:FSum}
\nabla_{\Tilde{\state}}\mathcal{F}
=  \nabla_{\Tilde{\state}}(\varepsilon_\obs^{\top}\Sigma^{-1}_{\ynoise}\bm \varepsilon_\obs) + \nabla_{\Tilde{\state}} (\varepsilon_\latent^{\top}\Sigma^{-1}_{\munoise}\bm \varepsilon_\latent)
\end{align}

\subsection{Confidence in estimation}
The inverse variance or precision ($\Sigma^{-1}=\!\Pi$) of the state estimate represents the agent's confidence in estimation. This precision also minimizes the free energy, simplified as the negative curvature of the internal energy at the estimated states \cite{friston2008variational,e23101306}:
\small
\begin{equation}
\begin{split}
            {\Pi}_\latent  = - \nabla_{\latent\latent} \mathcal{F}  = & - \big( D- \nabla_{\latent} f(\latent,\causes) \big)^T \Pi_\munoise  \big( D- \nabla_{\latent} f(\latent,\causes) \big) \\
      & - \nabla_{\latent}g(\latent)^T \Pi_\ynoise \nabla_{\latent}g(\latent),
\end{split}
\label{eqn:state_precision}
\end{equation}
\normalsize
where the generalized precision matrices ($\Pi_\munoise,\Pi_\ynoise$) are computed as given in Appendix \ref{appendix:gen-coords}.


AIF in continuous-time (with generalized coordinates) enables us to track the evolution of the probability density of the trajectory of states $\Tilde{\state}(t)$ \cite{Friston2008DEM}, instead of just its point estimates, thereby endowing the method with an accurate state estimate\footnote{When controlling a real robotic system, higher-order derivatives beyond accelerations are problematic when the noise smoothness properties are unknown.}. The key advantage of this model is its ability to leverage the generalized coordinates to capture the noise smoothness in data \cite{meera2020free}. However, the information contained in the higher-order noise derivatives is less valuable for the estimation process. Therefore, the noise precision matrices in the VFE expression should be designed such that the prediction errors coming from higher-order derivatives should be weighed less than those coming from lower-order derivatives. This raises the importance of noise precision modeling~\cite{meera2020free}. One of the main challenges while using generalized coordinates for real robots is that the quality of estimation is highly sensitive to the assumed noise smoothness of the signal, especially for low noise smoothness \cite{ajith2021DEM_drone,bos2021free}. To resolve this issue, future research can focus on the smoothness estimation of the coloured noise. Although a few attempts have been made in providing theoretical guarantees of stability and convergence for the estimation involving generalized coordinates \cite{ajith2021convergence,meera2020free,e23101306}, there is a huge scope for the proofs for optimality guarantees.

\section{Control}
\label{sec:aif_control}
Here we extend the previous section introducing the control actions. This is, how the robot both estimates its state and computes the control actions by filtering the information from previous observations according to its internal model dynamics. This can be seen as a low-level control where the actions correct for external and internal perturbation.

AIF robots use the same objective function for both estimation and control: the VFE. Thereby, the agent not only updates its state influenced by the world but can also apply actions to change the state of the world. This happens as an indirect consequence of actively sampling sensory data that is more in line with what is predicted by the internal model. Actions in active inference play a fundamental role in the approximation of the real distribution, acting also in the marginal likelihood by changing the real robot's configuration and modifying the sensory input $\obs$. The AIF robot is driven by:

\begin{subequations}
\begin{equation}
   \boldsymbol{\dot{\Tilde{\state}}} =  D\Tilde{\state} - \kappa_{x}\nabla_{\latent}\mathcal{F}(\latent, \obs)
\end{equation}
\begin{equation}
    \dot{\action} = -\kappa_u \sum_\obs \nabla_{\action} \obs \cdot \nabla_{\obs}\mathcal{F}(\latent, \obs)
    \label{eq:action_general}
\end{equation}
\end{subequations}

In AIF the control actions $\action$ steer the system towards minimizing the prediction errors in $\mathcal{F}$. This is again achieved through gradient descent. However, the VFE described is not a function of the control actions directly, but the actions $\action$ can influence $\mathcal{F}$ by modifying the sensory input. Thus, we can differentiate, in Eq.~\ref{eq:action_general}, observations w.r.t actions using the chain rule.  $\kappa_x$ and $\kappa_u$ are tuning parameters that define the step size in the iterative update. The partial derivatives of the sensory input with respect to the control action is a central point in active inference which has been tackled in different ways in past work like \cite{oliver2021empirical, pezzato2020novel}.

Conversely to other control frameworks, the desired state (goal or reference) is encoded in the internal state dynamics---Eq. (\ref{eq:statespace:dynamics})---as a preference. Thus, being away from the desired state increases the prediction error hence, generating the control actions that steer the system towards the goal. The properties and limitations of this formulation where $\mathcal{F}$ does not depend explicitly on the actions $\action$ are analysed in the discussion, as well as possible alternatives.



\section{Planning}
\label{sec:aif_planning}
In previous sections, we considered AIF in the context of continuous-time systems, where it was been cast as a gradient descent on instantaneous variational free energy (VFE). These sections showed that active inference can naturally describe estimation and control using a common probabilistic framework. In this section, we show how active inference can be used to explicitly model future states and observations in an action plan dependent manner, thereby enabling \emph{prospective planning}. This capability is crucial for many tasks and environments where actions have delayed consequences, where simply doing what is best in the current moment is not necessarily what is best in the long run.

To this end, we first introduce the concept of Partially Observable Markov decision process (POMDP) that we omitted in the introduction for clarity and model the control actions as a discrete-time optimization problem. Second, we define the expected free energy of the future (EFE) and finally, we describe the AIF approach to find the optimal control plan.

\subsection{Discrete-time optimization under the Markov assumption}
Modelling states and observations using generalized coordinates---as described in Sec.~\ref{sec:aif_estimation}---does allow the system to incorporate knowledge of the future, since generalized coordinates are a Taylor expansion in time around some given time point. However, to use generalized coordinates in practice, it is necessary to truncate them at some small order, meaning they can only model smooth and local changes over time. This is not sufficient for many complex control tasks where the consequences of action may occur far into the future. While there are alternative continuous-time methods to model future trajectories, these typically require substantially more mathematical machinery for limited algorithmic gain. Therefore, for reasons of mathematical tractability and computational (i.e., statistical) efficiency, AIF agents plans are typically constructed in discrete time. Moreover, it is further assumed that the relationship between states, observations and time are described by a POMDP \cite{kaelbling1998planning}. Intuitively, a POMDP assumes that there are discrete-time sequences of observations $\obs_{1:T}$ and hidden states $\state_{1:T}$\footnote{Again, for reasons of conceptual simplicity and mathematical tractability, we only consider future trajectories up to some time horizon $T$. Many of the results presented may apply in the infinite horizon case $T \rightarrow \infty$, but require more nuanced mathematical machinery to demonstrate.}. It is then assumed that, at some given instant point $k$, observations $\obs_k$ depend only on the hidden state $\state_k$. Similarly, it is assumed that the hidden states at the current time depend only on the hidden states at the previous time-step $\state_{k-1}$ (Markov assumption) and the action at the previous time-step $\action_{k-1}$. Analogously to the continuous-time space state AIF approach described in the previous sections, only current observation and previous state are needed to infer states and actions, instead of the entire history of states and observations, improving the tractability of the control problem. Additionally, these assumptions are not as restrictive in terms of generality as they first appear since the states are hidden, they can, by definition, include whatever information is necessary to ensure that the state at some instant $k$ depends only on the state and action at the previous time-step. 


The dependency structure entailed by the planning over time for stochastic variables is as follows,
\small
\begin{align}
    &p(\obs_{k:T}, \state_{k:T}, \action_{k:T}) = \nonumber\\
    &p(\obs_k | \state_k)p(\state_k)p(\action_k) \!\!\!\prod_{t=k+1}^T\!\!\! p(\obs_t | \state_t)p(\obs_k | \state_{k-1}, \action_{t-1})p(\action_t | \state_t)
\end{align}
\normalsize
This factorization of the joint distribution over observations, hidden states, and action trajectories will also mirror the structure of our active inference agent's generative model.

\subsection{Expected Free Energy}
To augment active inference for this scenario we introduce the concept of the \emph{expected} free energy (EFE)~\cite{friston2015active,parr2019generalised,millidge2021whence}. The EFE quantifies the average free energy of a plan-conditioned trajectory of states and observations, rather than the instantaneous free energy. Like instantaneous VFE agents are then mandated to minimize this quantity through both perception and action. An important terminological note concerns the word ``\emph{plan}" and its relation to the word policy. A plan is a sequence of actions $\pi = [\action_k, \action_{k+1} \dots \action_T]$. This differs from the policy used in reinforcement learning that refers to a parameterised action distribution conditioned on the current state. Interestingly, these control plans naturally confer active inference agents with both reward seeking and exploratory behaviours.


The EFE provides a measure quantifying the notion of the free energy expected over future trajectories. In effect, the EFE computes the average free energy of a trajectory, taking into account the the fact that because future observations are unknown they must be considered as random variables to be inferred, given a specific plan $\pi$.  Mathematically, the EFE ($\mathcal{G}$) is defined as,
\begin{align}
    \mathcal{G}_\pi = \mathbb{E}_{q(\obs_{k:T}, \state_{k:T} | \pi)}[\ln q(\state_{k:T} | \pi) - \ln \tilde{p}(\obs_{k:T}, \state_{k:T})]
\end{align}
Like the VFE introduced in previous sections, the EFE quantifies the difference between a variational density $q(\state_{k:T})$ and a generative model $p(\obs_{k:T}, \state_{k:T})$. However, the distributions are over trajectories of states and observations, instead of just the state and observation at a single time-step, and there is an additional expectation over future observations, meaning that EFE is a \emph{functional} of both states and observations, as opposed to only states.  In the continuous-time formulation, goals are encoded as set-points which, when compared against the current state estimates, form a prediction error that is minimized by action. In  POMDP formalism of EFE agents, the goals or `preferences` of the agent are encoded into the generative model to form a \emph{biased generative model} $\tilde{p}(\obs_{k:T}, \state_{k:T})$, where the tilde notation $\tilde{p}$ denotes that the model is biased towards predicting the agent's preferred environment, or equivalent, rewarding states and observations. 




\subsection{Finding the optimal plan}
One advantage of the EFE is that it allows the derivation of an expression for the optimal plan over a trajectory in terms of the sum of the EFE's for each individual time-step. This is possible due to the statistical factorizations intrinsic to the definition of the POMDP, whereby the current state  only depends on action and the state at the previous timestep. Specifically,  assuming the variational density also factorizes in time so that $q(x_{k:T}) = \prod_{t=k}^T q(x_t)$,  the EFE of a some trajectory can be decomposed into a  sum of the EFE for each individual time-step,
\begin{align}
\label{eq:1}
    \mathcal{G}(\obs_{k:T}, \state_{k:T}) = \sum_{t=k}^T \mathcal{G}(\obs_t, \state_t)
\end{align}
where,
\begin{align}
    \mathcal{G}(\obs_k, x_k) = \mathbb{E}_{q(\obs_k, \state_k)}[\ln q(\state_k) - \ln \tilde{p}(\obs_k, \state_k)]
\end{align}
Hence, the optimal plan (where a plan $\pi$ denotes a sequence of individual actions up to some time horizon $T$) is a softmax distribution over the sums of the EFE for each plan-conditioned trajectory. Specifically,

    \begin{align}
      \mathcal{G}(\pi) =&  \mathbb{E}_{q(\state_{k:T}, \obs_{k:T}, \pi)}[\ln q(\state_{k:T}, \pi) - \ln \tilde{p}(\obs_{k:T}, \state_{k:T}, \pi)] \nonumber\\
        =&\KL[q(\pi)||p(\pi) \sum_k^T - \mathbb{E}_{q(\obs_k, \state_k | \pi)}[\ln q(\state_k | \pi)  \\
        &- \ln \tilde{p}(\obs_k, \state_k | \pi)]] \nonumber\\
        =& \KL[q(\pi)||p(\pi) \sum_k^T -\mathcal{G}_{\pi_k}] \nonumber
         \end{align}
        The optimal posterior is given by is $\underset{\pi}{\mathrm{argmin}} \, \mathcal{G}(\pi)$
         \begin{align}
        &\implies q^*(\pi) = \sigma\left(\ln p(\pi) - \sum_k^T \mathcal{G}_{\pi_k}\right)\nonumber
    \end{align}

To gain an understanding of the kinds of behaviours that active inference agents will exhibit in practice, it is worthwhile studying the structure of the EFE objective in more detail. Crucially the EFE can be decomposed into two terms: an \emph{extrinsic} value term, which scores how close the agent is to achieve its goals, or to maximizing its reward or utility, and an \emph{intrinsic} value term which scores the information gain an agent could receive executing some plan. This information gain, mathematically, is simply the distinction between the posterior and prior variational distribution (or the agents' `beliefs') about the state trajectory---and maximizing this divergence essentially mandates the agent to seek out states which are maximally informative, thereby maximally reducing uncertainty. In effect, active inference agents possess a natural and inherent desire to seek out and explore novel states of the world which will cause them to update their world model. 

The EFE can be decomposed as follows---See Appendix \ref{appendix:EFE} for derivation,
\small
\begin{align}
    \mathcal{G}_\pi 
    \approx \underbrace{-\mathbb{E}_{q(\obs_k, \state_k | \pi)}[\ln \tilde{p}(\obs_k)]}_{\text{Extrinsic Value}} - \underbrace{\mathbb{E}_{q(\obs_k | \pi)}D_\mathrm{KL}[q(\state_k | \obs_k) || q(\state_k | \pi)]}_{\text{Intrinsic Value}}
\end{align}
\normalsize
which involves both \emph{extrinsic} and \emph{intrinsic} value terms. The exploratory, information-seeking behaviour induced by the intrinsic value term has been extensively studied in the active inference literature \cite{friston2015active,friston2017active} for its relationship to human notion of curiosity and intrinsic motivation that produces exploratory behaviours. Artificial curiosity~\cite{schmidhuber2010formal}, intrinsic motivation~\cite{oudeyer2009intrinsic} and goal-directed exploration~\cite{schwartenbeck2019computational} is crucial for learning and planning.
Furthermore, it has also been applied productively in reward-based AIF~\cite{millidge2020deep,tschantz2020reinforcement,van_der_Himst_2020,fountas2020deep} to handle agents which proactively learn to explore large state spaces with sparse rewards. 






\section{Learning}
\label{aif:learning}
In this section, we summarize how we can learn the generative models used in previous sections without the need of an expert designer. More importantly, function approximations~\cite{lanillos2018active,sancaktar2020end,ueltzhoffer2018deep,millidge2020deep} aid the extension of proposed framework to computationally tractable active inference agents that use high-dimensional inputs and states-spaces.

We can learn the likelihood $p(\obs_k | \state_k)$ and transition model (or prior) of the generative model---depending on the formulation $p(\state_k | \state_{k-1})$ or $p(\state_k | \state_{k-1}, \action_{k-1})$; the variational approximate posterior $q(\state_k | \obs_k)$, and the desired observation distribution $\tilde{p}(\obs_k)$. In the literature, two main approaches have been taken to achieve this: 1) discrete-state-space~\cite{friston2017active,da2020active} (Sec. \ref{sec:scaling:discrete}) and 2) \textit{function approximation} for adaptive control~\cite{lanillos2018adaptive,sancaktar2020end,meo2021multimodal} (Sec. \ref{sec:scaling:control}) and planning schemes~\cite{millidge2020deep} (Sec. \ref{sec:scaling:planning}).

\subsection{Discrete-state-space}
\label{sec:scaling:discrete}
Distributions are described as discrete categorical distributions, which explicitly enumerate every possible state and explicitly assign a probability to each. In practice, these distributions are implemented through normalized matrices and vectors representing each state, or each state combination. For instance, the likelihood mapping $p(\obs_k | \state_k)$ can be represented by an $\mathcal{Y} \times \mathcal{X}$ dimensional matrix where $\mathcal{Y}$ is the dimensionality of the observation space and $\mathcal{X}$ is the dimensionality of the action space.
Moreover, in a discrete state space setting, it is often tractable to explicitly evaluate the integral over time and compute the optimal plan posterior since agents in discrete states typically exist in small enough environments such that all policies can be explicitly enumerated and evaluated \cite{friston2017active}. Discrete state space active inference has been widely applied in computational neuroscience to simulate choice behaviour, e.g., saccades \cite{parr2021generative} and exploratory behaviour \cite{friston2015active}.

Discrete state space active inference suffers from clear limitations of scalability and expressiveness. First, the restriction of using discrete categorical distributions means that it must be possible to model the world with a discrete and low-dimensional set of states (to be able to successfully store the full matrices on a digital computer)\footnote{This problem may be addressed by having a hierarchy of states or by using sparse matrices where applicable. We leave it to future work to see whether this would enable discrete-state space active inference to scale to real-world tasks.}. This approach renders operating in any kind of continuous environment with fine-grained discretisation unfeasible. Second, and more importantly, explicitly evaluating the path integral in Equation \ref{eq:1} by enumerating all policies is an operation with exponential computational complexity in the time-horizon as well as the size of the state-space, since this increases with the branching factor of the policy tree. This exponential complexity very rapidly limits the scalability of this direct form of discrete state space active inference to relatively simple tasks such as the T-maze \cite{friston2015active}, although more recent studies are pushing the limit on the computational power of this method \cite{sajid2021active}.

\subsection{Deep active inference}
Another approach, which is substantially more scalable, although at the cost of losing performance and convergence guarantees, as well as interpretability, is to instead parametrize the distributions with general function approximators~\cite{lanillos2018active}, such as Gaussian processes~\cite{lanillos2018adaptive} or artificial neural networks~\cite{ueltzhoffer2018deep}. This allows, in theory, for \emph{any} distribution to be represented faithfully since deep artificial neural networks (ANN) can approximate any arbitrary nonlinear function given sufficient depth and width. For instance, the following approximate posterior
\begin{align}
q_\phi(\state_k | \obs_k) &= \mathcal{N}(\state_k; \mu_\phi(\obs_k), \sigma_\phi(\obs_k))
\end{align}
can be parametrized as follows,
\begin{align}
[\mu_\phi(\obs_k), \sigma_\phi(\obs_k)] &= f_\phi(\obs_k) 
\end{align}
where $f_\phi(\obs_t)$ represents the prediction (forward pass in a deep ANN with weights $\phi$) which outputs a predicted mean and variance for the Gaussian posterior as function of the input $\obs_t$. The parameters $\phi$ can be straightforwardly optimized, for instance using NNs, with the backpropagation algorithm where the loss function is the VFE or the EFE.

When the functions are approximated by deep nets the term coined is \textbf{deep active inference} (deep AIF). Here, first we describe the learning for adaptive control and second we explain how to use learning in planning. Furthermore, in the next Sec.~\ref{sec:aif_learning_hier} hierarchical learning is detailed.

\subsubsection{Adaptive control with deep AIF}
\label{sec:scaling:control}
Estimation and adaptive control can be scaled to high-dimensional inputs through stochastic optimization of the VFE exploiting deep ANNs~\cite{lanillos2021neuroscience}. Algorithm~\ref{algo:dai_control} describes the non-hierachical deep AIF for adaptive control. The most common approach is to learn the forward model $q_\phi(\latent)$---the sensory output $\obs_k$ given the internal state $\latent_k$---with self-supervised learning and then use the VFE optimization for the online estimation and control~\cite{sancaktar2020end,meo2021multimodal}. The state at instant $k$ is inferred using the forward pass of the network to get the estimated sensory input and backpropagating the weighted error (exploiting the Jacobian of the network, $\nabla_\latent q_\phi$). The action (e.g., torques, velocities, etc.) is computed with an analogous procedure but in this case the change of the sensory input w.r.t the action should be modeled $\nabla_\obs \action$. There are several techniques to resolve this partial derivative~\cite{oliver2021empirical, pezzato2020active}. This approach can be further extended to multimodal sensory input~\cite{meo2021multimodal} and combined with model agnostic AIFs approaches.

\begin{algorithm}[hbtp!]
	\caption{Deep active inference for adaptive control}\label{algo:dai_control}
	\begin{algorithmic}[1]
		\small
		\State Generative model $p( \state_k, \obs_k \phi)$
		\State Approximate posterior $q(\state_k, \phi)$
		\While{true}
		\State $\obs_k \gets \mathrm{env.step(\action)}$ \Comment{Update environment given action}
		\State $\epsilon_{\obs_k} \gets \Pi_\obs (\obs_k- q_\phi(\latent_k))$ \Comment{Prediction error obs.}
		\State $\epsilon_{\latent_k} \gets \Pi_\latent (D\latent_k- f_\phi(\latent_k,\causes_k))$ \Comment{Prediction error dyn.}
		\State $q(\state_k)\equiv \latent_k \gets D\latent +\nabla_\latent q_\phi\epsilon_{\obs_k} + +\nabla_\latent f_\phi\epsilon_{\latent_k}$ \Comment{Perception}
		\State $\action \gets -\nabla_\obs \action \nabla_\latent q_\phi\epsilon_{\obs_k}$ \Comment{Control}	
		\EndWhile
	\end{algorithmic}
\end{algorithm}
\normalsize

\subsubsection{Planning with deep AIF}
\label{sec:scaling:planning}
Planning ahead can be modeled with deep AIF. Here all densities (even the policy) and parameters are learnt by optimizing the EFE. The algorithm is described in Alg.~\ref{algo:dai}. This deep AIF approach draws heavily from recent work in machine learning, especially deep RL, and has achieved significant successes at enabling active inference to be scaled up to challenging RL benchmark tasks, and to meet, and potentially exceed the state of the art in deep RL \cite{millidge2020deep,tschantz2020reinforcement, tschantz2020scaling,van_der_Himst_2020}.

Interestingly, if we explicitly model the parameters in the generative and approximate posterior distributions, the EFE objective also gives rise to an information gain term over the model parameters, which encourages deliberate exploration to efficiently search the parameter space. The ensuing exploration is usually described as one aspect of artificial curiosity or intrinsic motivation~\cite{schmidhuber2010formal,oudeyer2009intrinsic,schwartenbeck2019computational}; namely, novelty seeking.

While this approach allows for the expression of arbitrary probability densities, it does not solve the problem of the computationally expensive evaluation of the EFE path integral, used to compute the optimal policy. However, in the deep AIF literature there have been several proposed solutions which draw either from model-based or model-free RL. One approach is to use the fact that the EFE satisfies a similar recursive relationship as the Bellman equation in traditional RL to derive a bootstrapped estimator similar to a value function, which can be optimized across batches \cite{millidge2020deep}. Another approach is to utilize black-box optimization methods, such as genetic algorithms, to directly approximate this quantity through sampling \cite{ueltzhoffer2018deep}. A final approach, which is closely related to model-based planning, and is more common in the literature, is to approximate the path integral with samples of future roll-outs given different policies, which can be simulated by the agent using its own generative model to imagine and score future trajectories \cite{tschantz2020reinforcement}. This model-based approach has been used fairly widely in the literature, and has been utilized to develop powerful deep active inference algorithms capable of matching the performance of state of the art model-based RL systems in MDP problems, such as the Cartpole, Acrobot, Lunar-lander~\cite{millidge2020deep}, the mountain car problem~\cite{catal2019bayesian}; and POMDPS as visual control of the cartpole~\cite{van_der_Himst_2020}, the car racing~\cite{ccatal2020learning,van2021deep} and the Animal-AI environment~\cite{fountas2020deep}.

\begin{algorithm}[hbtp!]
  \caption{Deep active inference for planning}\label{algo:dai}
  \begin{algorithmic}[1]
  	\small
      \State Generative model $p(\obs_k, \state_k, \pi, \phi)$
      \State Approximate posterior $q(\state_k, \pi, \phi)$
      \While{$k \leq T $}
        \State $\obs_k = \mathrm{env.step(\pi)}$ \Comment{Update environment given policy}
        \State $q(\state_k| \pi) \gets \arg\min_{q(\state_k|\pi)} \ \mathcal{F}(q, \obs_k)$ \Comment{Perception}
        \State $q(\phi) \gets \mathrm{argmin}_{q(\phi)}  \ \mathcal{F}(q, \obs_k)$ \Comment{Parameters learning}
        \State $q(\pi) \gets \mathrm{argmin}_{q(\pi)} \ \mathcal{G}(q, \obs_k)$ \Comment{Control}
        \State $\pi \sim q(\pi)$ \Comment{Select most likely policy}
      \EndWhile\label{euclidendwhile}
      \State \textbf{return} $q(\state_k, \pi, \phi)$
  \end{algorithmic}
\end{algorithm}
\normalsize

\section{Hierarchical Representation}
\label{sec:aif_learning_hier}
In previous sections, we described the state in AIF as a variational density. This section introduces how to represent the state of the robot in a hierarchical setting. Using a hierarchical recurrent neural network (RNN) we can generate predictive behaviour exploiting the free energy principle. 
Murata and colleagues~\cite{murata-tani2015} developed a variational hierarchically-organized RNN model, referred to as the Stochastic Continuous-Time RNN (S-CTRNN). 
This model inherits the dynamic property of the multiple timescales RNN (MTRNN)~\cite{yamashita-tani2008} wherein the neural activation dynamics in the higher layer is dominated by slower dynamics by employing a larger time constant and the one in the lower layer is dominated by faster dynamics with a smaller time constant. 
Although this model was successfully applied for predictive coding and active inference of a physical humanoid robot that imitatively interacts with a human-operated robot, the model was limited since the random latent variable is allocated only in the initial latent state of the MTRNN.

\begin{figure}[hbtp!]
	\centering
    \includegraphics[width=1.0\columnwidth, height=170px]{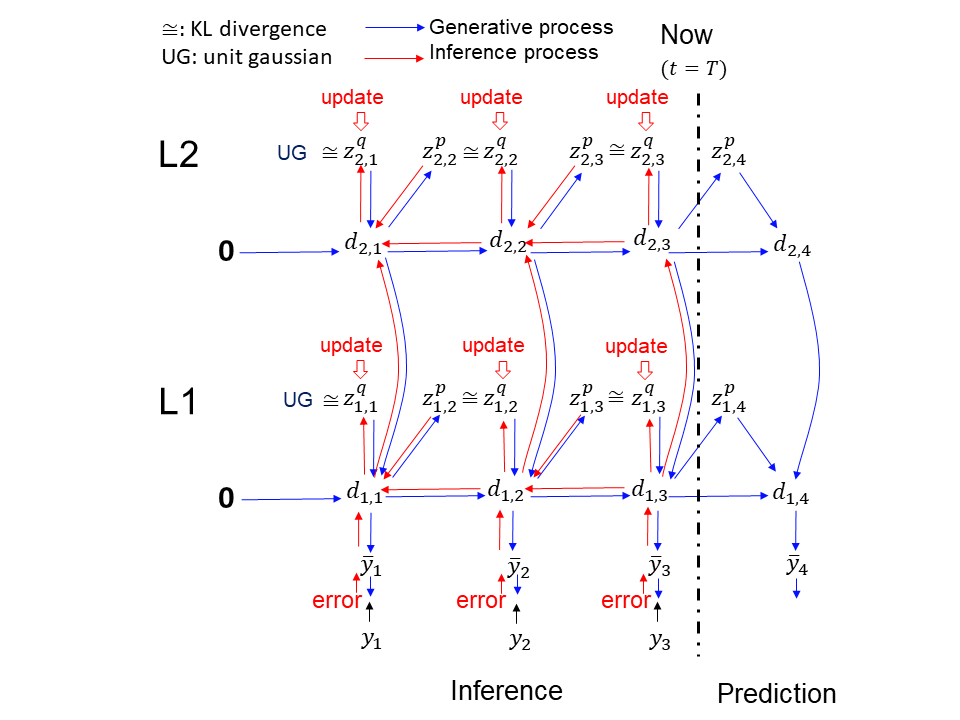}
	\caption{\textbf{PV-RNN graphic diagram}. It illustrates a case of action generation wherein the inference is performed for the past window from $k=1$ to $k=3$ and the prediction for the future time step $k=4$ is performed.}
	\label{fig:III-H_PV-RNN}
\end{figure}

By considering that random latent variables should be introduced not only in the initial time step but also all time steps during the sequential processing, another variational RNN, so-called the predictive coding inspired variational RNN (PV-RNN) \cite{ahmadi-tani2019} was developed. 
This model was inspired by the sequence prior scheme \cite{chung2015recurrent} evinced in a model called the variational RNN (VRNN) wherein the prior changes over time.
PV-RNN using the sequence prior was extended incorporating (1) a scheme called error regression to infer random latent variables using prediction error signals during action generation which is analogous to predictive coding (2) the multiple timescale scheme described above.
A graphical representation of PV-RNN is shown in Figure~\ref{fig:III-H_PV-RNN}.
In each layer $l$ of the network we define the internal state $\latent$ with two latent variables: deterministic $\bm{d}_k^l$ and random $\bm{z}_k^l$, respectively with $k$ as time step. The random variable $\bm{z}_k^l$ is represented by a Gaussian distribution with mean $\mu$ and standard distribution $\sigma$.
The lowest layer contains the output $\obs_k^l$ predicting sensory observation including both exteroception and proprioception. In robotics applications, motor movement can be generated by feeding the prediction of proprioception in terms of target joint angles in the next time step to the motor controller.

Both learning and action generation is performed through iterative interactions between the top-down generative process and the bottom-up inference process.
The generative model can be written as factorized:
\small
\begin{equation}
\begin{aligned}
\label{eq:genp} 
    p({\obs}_{1:T}, {\mathbf{d}}_{1:T}, \bm{z}_{1:T} | \bm{d}_0) = 
    \prod_{k=1}^T p({\obs}_k | {\mathbf{d}}_k)  p({\mathbf{d}}_k | {\mathbf{d}}_{k-1}, \bm{z}_k) p(\bm{z}_k | {\mathbf{d}}_{k-1})
\end{aligned} 
\end{equation}
\normalsize
Although $\mathbf{d}$ is a deterministic latent variable, it can be considered to have a Dirac delta distribution centered on $\tilde{\bm{d}}$ as $\bm{\sigma}(\bm{d}-{\tilde{\bm{d}}})$.
In the generative process, the information propagates from the highest layer to the lowest layer at each time step in the forward direction. At each layer the deterministic latent variable $\v{\tilde{d}}_k^l$ and the prior distribution $p(\v{z}_k^l)$ are computed. The lowest layer computes the sensory prediction $\obs_k^l$.
First, $\v{\tilde{d}}$ is computed as shown in Eq. (\ref{eq:gene-d}) where its internal value is denoted by $\bm{h}$. 
\small
\begin{equation} \label{eq:gene-d}
\begin{aligned} 
    \v{\tilde{d}}^l_k &= \text{tanh}(\bm{h}^l_k)\\
    \bm{h}^l_k &= \left(1 - \frac{1}{\tau^l}\right)\bm{h}^l_{k-1} 
    \\&+ \frac{1}{\tau^l} \left( \bm{W}^{l,l}_{d,d}\bm{\tilde{d}}^l_{k-1} + \bm{W}^{l,l}_{z,d}\bm{z}^l_k + \bm{W}^{l+1,l}_{d_u ,d}\bm{\tilde{d}}^{l+1}_{k} + \bm{b}^l_k \right)
\end{aligned}
\end{equation}\normalsize
where $\v{z}^l_k$ is a sampling of the posterior distribution $q(\v{z}_k^l)$ estimated in the last iteration of the inference process.
$\bm{W}$ represent the learnable parameters in terms of connectivity weight matrices between layers and their deterministic and stochastic units. $\bm{b}$ represents another learnable parameter bias.
The output is computed as mapping from $\bm{\tilde{d}}^{1}$ in the lowest layer as:
\begin{equation} \label{eq:output} 
\begin{aligned}
\mathbf{\bar{y}}_k = \bm{W}^{ll}_{yd}\bm{\tilde{d}}_k^1 + \bm{b}_y
\end{aligned}
\end{equation}
where $p(\v{b}_y)$ is a learnable bias.
\par
The prior distribution $p(\v{z}_k)$ takes a Gaussian distribution with mean $\bm{\mu}_k^p$ and standard deviation $\bm{\sigma}_k^p$. The prior depends on $\bm{\tilde{d}}_{k-1}$ by following the sequence prior scheme \cite{chung2015recurrent}.
\begin{equation} \label{eq:gene-p}
\begin{aligned}
    p(\v{z}_k | \bm{\tilde{d}}_{k-1}) &= \mathcal{N}(\bm{\mu}^p_k, (\bm{\sigma}^p_k)^2) \text{ where $k>1$}\\
    \bm{\mu}^p_k &= \text{tanh}(\bm{W}^{l,l}_{d,z,\mu^p}\v{\tilde{d}}_{k-1})\\ 
    \bm{\sigma}^p_k &= \exp(\bm{W}^{l,l}_{d,z,\sigma^p}\v{\tilde{d}}_{k-1})
\end{aligned}
\end{equation}

Next, the inference process using the free energy minimization is described. The VFE $\mathcal{F}$ can be written as:
\begin{equation}
    \mathcal{F}=-\underbrace{\mathbb{E}_{q_\phi(\boldsymbol{\state}|\boldsymbol{\obs})}[\ln p_\theta(\boldsymbol{\obs}|\boldsymbol{\state})]}_{\rm Accuracy}+\underbrace{\KL[q_\phi(\boldsymbol{\state}|\boldsymbol{\obs})\Vert p(\boldsymbol{\state})]}_{\rm Complexity}\label{eq:femin}
\end{equation}
where $p_\theta(\boldsymbol{\obs}|\boldsymbol{\state})$ is the generative model with learnable parameter $\theta$ and $q_\phi(\boldsymbol{\state}|\boldsymbol{\obs})$ is the posterior inference model with parameter $\phi$. 
The objective of the inference in learning is to obtain optimal values for $\theta$ and $\phi$ by minimizing the free energy.
Let us consider the posterior inference with a given sensory observation $\overline{\obs}_{0:T}$ for the past-time window from time step $k=1$ to $T$ of PV-RNN.
The approximate posterior for $\bm{z}_{0:T}$ is represented as:
\begin{equation} \label{eq:q}
\begin{aligned}
    q(\v{z}_k | \obs_{k:T}) &= \mathcal{N}(\bm{\mu}^q_k, \bm{\sigma}^q_k)\\
    \bm{\mu}^q_k &= \text{tanh}(\bm{A}^\mu_k)\\
    \bm{\sigma}^q_k &= \exp(\bm{A}^\sigma_k)
\end{aligned}
\end{equation}
The adaptation variable $\mathbf{A}_{0:T}$ represents the parameters $\phi$ for the posterior inference model $q_\phi$.
With considering the generative process described in Eq.~\ref{eq:gene-d} and Eq.~\ref{eq:gene-p}, the free energy $\mathcal{F(\phi, \theta)}$ for PV-RNN can be obtained as:
\begin{equation} 
\begin{aligned} 
    \mathcal{F}(\phi, \theta) 
    & = -\sum_{k=1}^T (E_{q_\phi(\v{z}_k | \obs_{k:T})} \big[\log  p_{\theta_x}(\bm{\obs}_k | \v{\tilde{d}}_{k}) \big] \\
    & + \sum_{l}^L \KL\big[ q_\phi(\v{z}_k | \obs_{k:T}) || p_{\theta_z} (\v{z}_k | \v{\tilde{d}}_{k-1}) \big])
\label{eq:feminRNN}
\end{aligned}
\end{equation}
where the first term is the accuracy and the second term is the complexity as corresponding to those in Eq.~\ref{eq:femin}.
In practice, the generative process---sweeping from the higher layer to the lower layer and from time step $0$ to $T$---is performed by following Eq.~\ref{eq:gene-d} and Eq.~\ref{eq:gene-p} using $\mathbf{A}_{1:T}$ and the connectivity weight $\bm{W}$ updated in the previous iteration. The inference process---sweeping from the lower layer to the higher layer and from time step $T$ to $1$---is implemented with back-propagation through time (BPTT) \cite{rumelhart1985learning}. This uses the error calculated between the output computed in the last generative process and the target for updating the current $\mathbf{A}_{1:T}$ and $\bm{W}$.
The paired computation for the generative process and the inference process is iterated until the free energy $\mathcal{F(\phi, \theta)}$ is minimized. After the learning process has converged, action generation can be conducted.


The basic architecture described above has been extended and implemented in various robotic experimental tasks including human-robot imitative interaction \cite{chameICRA2020,chameFrontier2020}, dyadic robot imitative interaction \cite{wirkuttis-tani2021}, goal-directed planning for robot object manipulation \cite{matsumoto2020goal}, and goal-directed planning using active inference and reinforcement learning in navigation \cite{Han-Tani-RL-AIF-2021}.

\section{Robotic experiments}
\label{sec:applications}
While the previous sections were devoted to describing the mathematical insights of AIF. This section presents selected experiments in the literature that showcase the relevant characteristics of AIF for robotics.\\
\\
\noindent \textbf{Historical preamble}. 
While Friston was developing the basis of AIF, the free energy principle~\cite{friston2010free}, Tani and colleagues \cite{tani2003, tani2004} were investigating models similar to AIF in real robots. In \cite{tani2003}, a hierarchically organized RNN as a generative model was trained to predict/generate visuo-proprioceptive sequence patterns for a set of movement primitives. It was demonstrated that the robot could successfully adapt its movement pattern to the corresponding movement primitive in real-time when the environment changed. However, these models are limited because they were predicated on deterministic dynamics perspective instead of a Bayesian perspective which is used in the formal formulation of AIF~\cite{friston-biol-2011}. A robotic attempt of AIF with its exact Friston's formalism for reaching tasks was described in~\cite{pio2016active}, with a 7-DOF simulated robot arm with the generative models and parameters known in advance. Finally, Lanillos and colleagues~\cite{lanillos2018adaptive,oliver2021empirical}, were able to develop and deploy an AIF model on a humanoid robot for estimation and control. Concurrently, similar approaches were being investigated for industrial manipulators~\cite{pezzato2020active}.


An important concept that emerged in these works is that actions in AIF realize sensory consequences of prior causes. For this reason, there is no need for precise inverse dynamical models which might be problematic to compute. While compared to optimal control, AIF is quite appealing for robotic applications where the dynamics of the robot or the task are uncertain. However, other challenges appear, e.g., inverse dynamic modelling is shifted to the design/learning of meaningful generative models and prior beliefs (preferences). Despite the challenges, AIF is showing huge potential for real robotic applications~\cite{lanillos2021neuroscience} for estimation, adaptive control, fault-tolerant control, prospective planning and complex cognition skills (i.e., human-robot collaboration, self/other distinction).

\begin{figure*}[hbtp!]
    \centering
    \includegraphics[width=0.95\textwidth, height=245px]{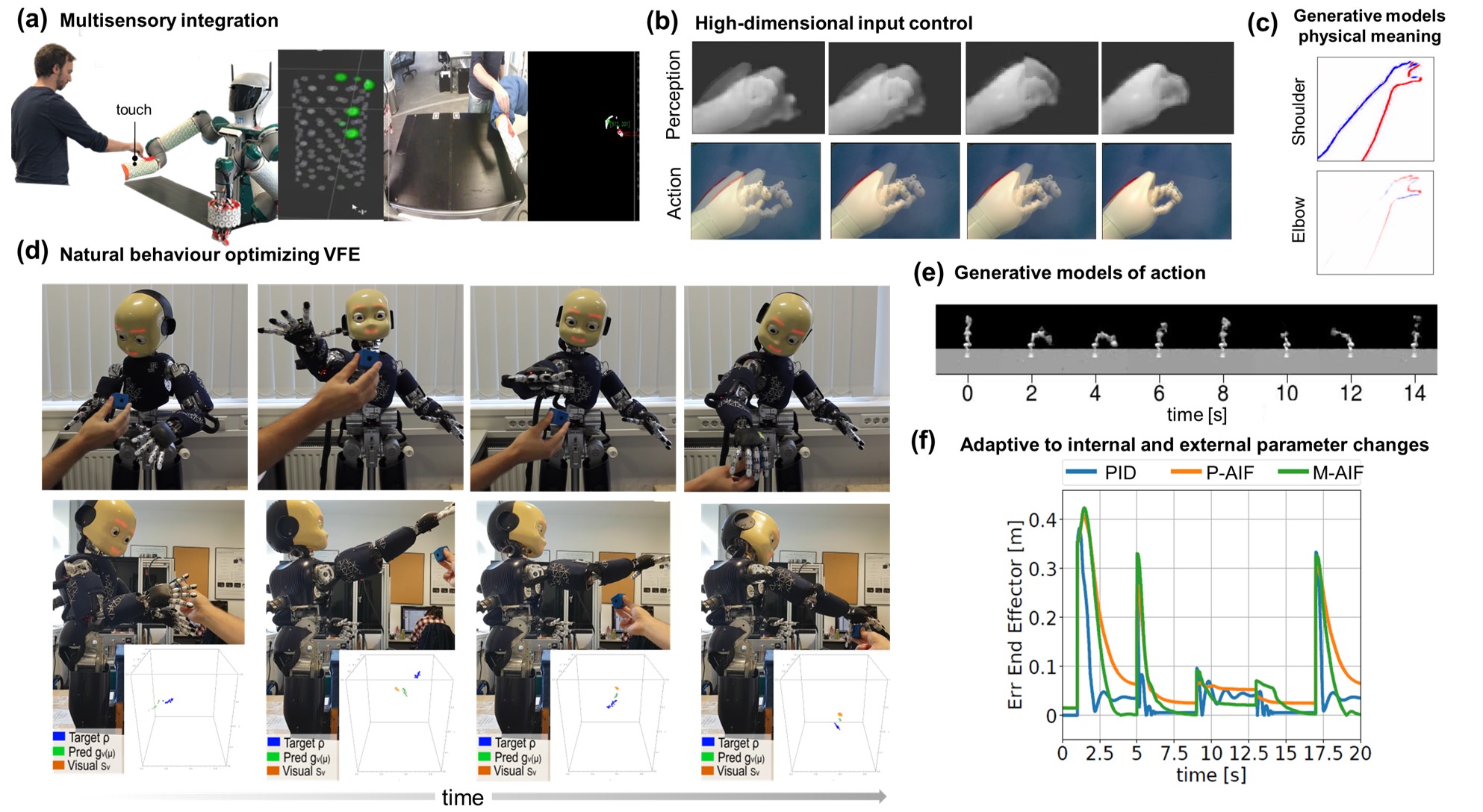}
    \caption{\textbf{AIF estimation and control benefits}. (a) Multisensory integration (proprioceptive, visual and tactile body estimation) adapted from~\cite{lanillos2018adaptive}. (b) High-dimensional input (images) state-estimation and control through generative model learning, taken from~\cite{sancaktar2020end}. (c) Generative models have a physical meaning (backward neural network pass (Jacobian) visualization w.r.t. the elbow and shoulder for a virtual arm model after learning, taken from \cite{rood2020deep}. (d) Natural behaviour through VFE minimization (upper-body reaching and head-eyes attention), taken from~\cite{oliver2021empirical}. (e) AIF provides generative models of action and permits trajectory imagination. (f) Adaptive to internal and external parameter changes (e.g., stiffness, gravity). The plot compares a builtin controller from the PANDA robot against the AIF using proprioceptive information (P-AIF, \cite{meo2021multimodal}) and multisensory integration (M-AIF), adapted from~\cite{meo2021multimodal}.} \label{fig:aif-lanillos}
\end{figure*}

In the following subsections we describe experiments for: A) estimation, B) adaptive control, C) fault-tolerant control, D) planning and E) complex cognition.

\subsection{Estimation}
VFE optimization has been successfully applied for adaptive robot estimation and learning in applications, such as body estimation~\cite{lanillos2018adaptive}, drones state tracking~\cite{meera2020free} and navigation~\cite{catal2021robot,burghardt2021robot}.

In~\cite{lanillos2018adaptive} the authors proposed a computational model based on predictive coding which allows \textbf{generic multisensory integration} for inferring and learning its body configuration by means of arbitrary sensors affected by Gaussian noise. Figure~\ref{fig:aif-lanillos}a shows the experimental setup. They learned the forward model using Gaussian processes. This is particularly useful in the case an accurate model of the body and the environment is not available, or when self-calibration of different sensors is needed. Body learning was formulated as obtaining the mapping that encodes the sensory value dependency on the joint variables. Body perception/estimation was achieved by minimizing the VFE (through stochastic gradient descent), which iteratively reduces the discrepancy between the belief of the robot about its configuration and the observed posterior. The results showed that different sensor modalities improve the refinement of the body configuration and allow the body estimation to adapt to the most plausible solution when injecting strong visual-tactile perturbation or in the case of missing sensory inputs. Interestingly, the system was prone to visual-tactile illusions, similarly to how humans process body multisensory information~\cite{hinz2018drifting}.

Based on the Dynamic Expectation Maximization (DEM) \cite{friston2008variational}, \cite{meera2020free} proposed a simultaneous state and input observer design for tackling stable Linear Time Invariant (LTI) systems with \textbf{coloured noise}, which was tested on a real system. The use of generalized coordinates enabled this observer to outperform the Kalman filter for state estimation and Unknown Input Observer (UIO) for input estimation, under coloured noise. Similar ideas of perception were used to reformulate DEM into a blind \textbf{system identification} algorithm \cite{e23101306} for the simultaneous estimation of states, inputs, parameters and noise hyperparameters of an LTI system. This estimator was shown to outperform other classical estimators for parameter estimation under coloured noise. On the application side, DEM was applied for the perception of a quadrotor flying under wind conditions \cite{ajith2021DEM_drone,bos2021free}. The DEM based learning algorithm successfully learned the dynamic model of the quadrotor for accurate output predictions when compared to other state-of-the-art system identification methods \cite{ajith2021DEM_drone}. The existence of a mathematical proof for stable parameter estimation \cite{ajith2021convergence} motivates its reliability and safety for real robotic applications. These results demonstrate the applicability of DEM as a learning algorithm for future robots to learn their generative model from sensory data, rendering them with the capability to make accurate predictions of the world. 

\subsection{Adaptive control}
\subsubsection{Without learning}
AIF robot body perception and action were successfully deployed, for the first time, on a physical robot~\cite{oliver2021empirical}---the iCub humanoid robot. It provided \textbf{natural behaviours} for upper body reaching and head object tracking using proprioceptive and visual sensory inputs. The AIF algorithm was validated in terms of noise robustness, and multisensory integration, and was also applied for reaching and grasping a moving object. Figure~\ref{fig:aif-lanillos}e shows the behavioural sequence for one of the tests. The chosen models for active inference allowed to express goals in Cartesian or image spaces by specifying attractors in this space and translating them into joint space by using the inverse of the Jacobian matrix. The robot was controlled using velocity commands which allow to easily define the partial derivatives of the joint position with respect to the control actions considering discrete updates of the joint positions at each time step. This removed the need for complex inverse dynamical models to compute the control actions. The experimental results showed the potential of AIF for dual perception and action, particularly for counteracting environmental or model unexpected changes.



AIF for \textbf{torque-control} in the joint space for industrial arms was shown in~\cite{pezzato2020novel}. The authors presented an adaptive scheme for controlling a generic $n$-DOFs robot manipulators, namely Active Inference Controller (AIC). The control law is model-free, lightweight, and it uses proprioceptive sensors for position and velocity to control a 7-DOF robot arm in joint space. By choosing the state of the system to be controlled as the joint positions, the generative models of the sensory input simply resulted in the identity mapping. The generative function of the state dynamics was used to impose a specific behaviour. The robot believed that each joint was moving towards a given target following the dynamics of a chosen first-order linear system. Furthermore, instead of computing the forward dynamics of the robot manipulator, the authors in~\cite{pezzato2020novel} proposed to approximate the partial derivatives of $\mathcal{F}$ with respect to the control input by just encoding the sign of this relationship, relying on the adaptability of the algorithm to compensate for the modelling errors. Note that by specifying a certain generative model in AIF, one can define a robot that \textit{thinks it will behave in a particular way}. This is closely related to the more established idea of model reference adaptive control (MRAC)~\cite{astrom}. The results showed that the AIC greatly outperformed the MRAC in terms of adaptability to unmodelled dynamics, tuning effort, disturbance rejection, computational effort, and overall performance in pick and place tasks with unknown object weights. A great advantage of this approach was the ability to transfer from simulation to real robot without re-tuning of the controller while preserving compliant behaviour through torque commands. The model parameters for the AIC have however been considered constants and they were seen as tuning parameters for the designer to achieve a smooth response. Also, the Mutisensory AIF presented in~\cite{meo2021multimodal} showed better performance than the optimized factory controller of the Panda robot (Fig.~\ref{fig:aif-lanillos}f) when \textbf{adapting to external and internal parameters changes}, such as gravity, stiffness, end-effector inertia and external perturbations.

\subsubsection{With precision learning}
The authors in \cite{baioumy2020active} presented an evolution of \cite{pezzato2020novel} by including online parameters learning for controller auto-tuning. The authors demonstrated how the minimization of the VFE produces effective state-estimation, control for robotic manipulators. They introduced a temporal parameter $\tau$ in the dynamics generative model used in \cite{pezzato2020novel} corresponding to a variable time constant of the first-order linear system used as the model reference for the joint motions. 
Basically, the authors showed that the gains of the controller correspond to the covariance matrices of the observation model, and that learning the optimal covariance matrices results in finding the optimal gains for the controller~\cite{baioumy2020active}. The results showed improved performance in terms of response and robustness compared to a manually tuned AIC, but they also showed some limitations of the standard active inference formulation when \mbox{(hyper-)parameters} learning is introduced. The belief about the current state is intentionally biased towards the target to achieve control, which is an uncommon thing in the control community. Thus, the state reconstruction is not accurate unless the system reaches the target. This is reflected also in the learned covariance matrices that converged to a much higher value than the Gaussian noise affecting the controlled system. 


\subsubsection{With function learning}
Instead of designing the observation and the dynamical model, a solution for body estimation and control was proposed in \cite{Lanillos2018ActiveIW}, where function learning was employed. In this approach, only forward models needed to be learned. The authors pointed out how this learning approach was not simpler than classical inverse dynamics techniques, because it required learning the state forward dynamics and the Jacobian of the observation model with respect to the latent space.

In~\cite{lanillos2018adaptive} AIF estimation with function learning was solved using GP regression when the input is low-dimensional. It exploited the closed-form equation to compute the partial derivatives with respect to the body state. A pixel-based deep AIF controller \cite{sancaktar2020end} was presented to handle \textbf{high-dimensional inputs}, such as images. Figure~\ref{fig:aif-lanillos}b depicts the Pixel-AIF architecture and a sequence of the dual perception-action inference process. In the perception row, the arm overlays represent the robot visual input and the predicted image. In the action row, the overlays represent the visual input and the desired goal in the visual space. This approach incorporated generative model learning using convolutional neural networks for one sensor modality (visual) and performed velocity control. Finally, a multimodal variational autoencoder AIF (M-AIF) torque controller \cite{meo2021multimodal} was presented. It combined VFE optimization with generative model learning, extending the previous AIF formulations to work with high-dimensional multimodal input at the torque level. Figure~\ref{fig:aif-lanillos}e shows the robot imaging its future trajectory thus, describing AIF as a \textbf{generative model of actions}. Figure~\ref{fig:aif-lanillos}f shows the M-AIF results with the panda manipulator robot when changing external parameters: the Jupiter experiment. The gravity parameter was modified to $24.79  {m}/{s^2}$. According to the end-effector error plot, the AIF formulation achieved better performance when compared with the built-in controller provided by the panda company. Further tests showed the adaptation properties to input noise and internal parameters changes, such as stiffness.

In all these methods the computation of the observation model Jacobian (inverse model) is indirect and depends on the proper representation learning. This Jacobian has a \textbf{physical meaning} and describes the change in perception or the action to be exerted. Figure~\ref{fig:aif-lanillos}c visualizes the convolutional decoder Jacobian of a virtual arm after learning the visual mapping~\cite{rood2020deep}. The red and blue pixels defines the change on the arm edge w.r.t. to the elbow and shoulder joints.

Other works focused on robot navigation with active inference. In particular, \cite{catal2020deep,ccatal2020learning,catal2021robot} proposed a method to learn generative state-space models from pixel data. In \cite{catal2020deep} the authors approximated the variational posterior distributions, as well as the likelihood model with two deep neural networks. This allows performing simple navigation tasks using a Kuka YouBot platform in an aisle. 

\subsubsection{In combination with discrete planning}
Adaptive control can also be achieved for high-level behaviour with active inference, as shown in \cite{pezzato2020activeBT}. A discrete formulation of active inference is used in combination with behaviour trees \cite{colledanchise2017} to provide prior preferences over specific states while adapting at runtime to unforeseen contingencies during mobile manipulation in dynamic environments. The solution proposed in \cite{pezzato2020activeBT} blended acting and planning to provide a solution for continual planning and hierarchical deliberation for long term tasks in robotics. The core idea is to use behaviour trees as a graphical method to encode priors for the active inference algorithm. These priors were used to bias the generative model online and to guide the search with active inference to provide adaptive and fast responses to changes in the environment. The experiments were conducted both in simulation and in a real robot considering two different mobile manipulators and two similar tasks in a retail environment, for instance stocking an empty shelf. The hybrid combination of active inference and behaviour trees provides reactivity to unforeseen events, allowing the mobile manipulators to quickly perform, repeat, or skip actions according to the state of the environment. Crucially, the method also provides safety guarantees and convergence of the high-level behaviour of the robot to the given goal.

\subsection{Fault-tolerant control}
AIF can also help in case of degraded sensory input if sensory redundancy is provided in the system~\cite{lanillos2018adaptive}. As pointed out in \cite{pio2016active}, if the robot has poor proprioceptive information, multimodal integration with visual data can compensate and restore effective control~\cite{lanillos2018adaptive}. This property of active inference of naturally fusing different sensory modalities for both state estimation and control has been taken further for \textbf{fault-tolerant control}~\cite{pezzato2020active, baioumy2021fault}.

The work in \cite{pezzato2020active} proposes the use of the AIC from \cite{pezzato2020novel} with the addition of visual information on the end-effector position---using the GP learning approach from~\cite{lanillos2018adaptive}. The authors proposed a scheme for online threshold generation for fault detection and isolation of sensory faults based on the sensory prediction errors in the free-energy. Fault recovery when either the proprioceptive or the camera were marked as faulty is achieved by simply setting to zero the precision matrices (or inverse covariances) of the relative sensors. Results on a simulated 2-DOF robot arm showed that the fault-tolerant AIC can detect and recover from freezing encoders and camera misalignment providing convergence to a given goal. The main advantage with respect to standard fault-tolerant approaches is that fault detection and isolation, as well as fault recovery, do not require the design of additional signals to monitor or alternative controllers besides what is already provided by the AIC. However, as AIF is by nature an adaptation mechanism biased towards the desired state it hindered fault detection, producing false alarms. In fact, the sensory prediction errors in the free-energy can increase due to a changing goal for the robot manipulator and not necessarily due to a faulty sensor. This problem can be addressed by introducing an unbiased AIC controller~\cite{baioumy2021fault}. 

Subsequently, the authors in \cite{baioumyIWAI2021} highlighted how fault detection and recovery can be automatically achieved through precision learning. This provides a method for stochastic fault-detection (the probability of sensory being fault) rather than deterministic and allowed for fault-tolerant behaviour without needing any threshold definition.


\subsection{Planning}
\label{sec:sota:planning}

As described in section \ref{sec:aif_planning}, planning can be cast as optimizing the parameters of the plan density $q(\pi)$ with respect to the cost function EFE $\mathcal{G}(\pi)$. This process favors plans which realise an agent's prior preferences (i.e. goal-driven behaviour), while at the same time gathering the most information from the environment (i.e exploration).

A classical illustrative example is modelling saccadic eye movements. In \cite{friston2017graphical} they computed the saccades (using expected free energy, EFE) and then executed them through vanilla prediction error minimization (free energy gradients).

\begin{figure}[hbtp!]
    \centering
    \includegraphics[width=0.85\columnwidth, height=180px]{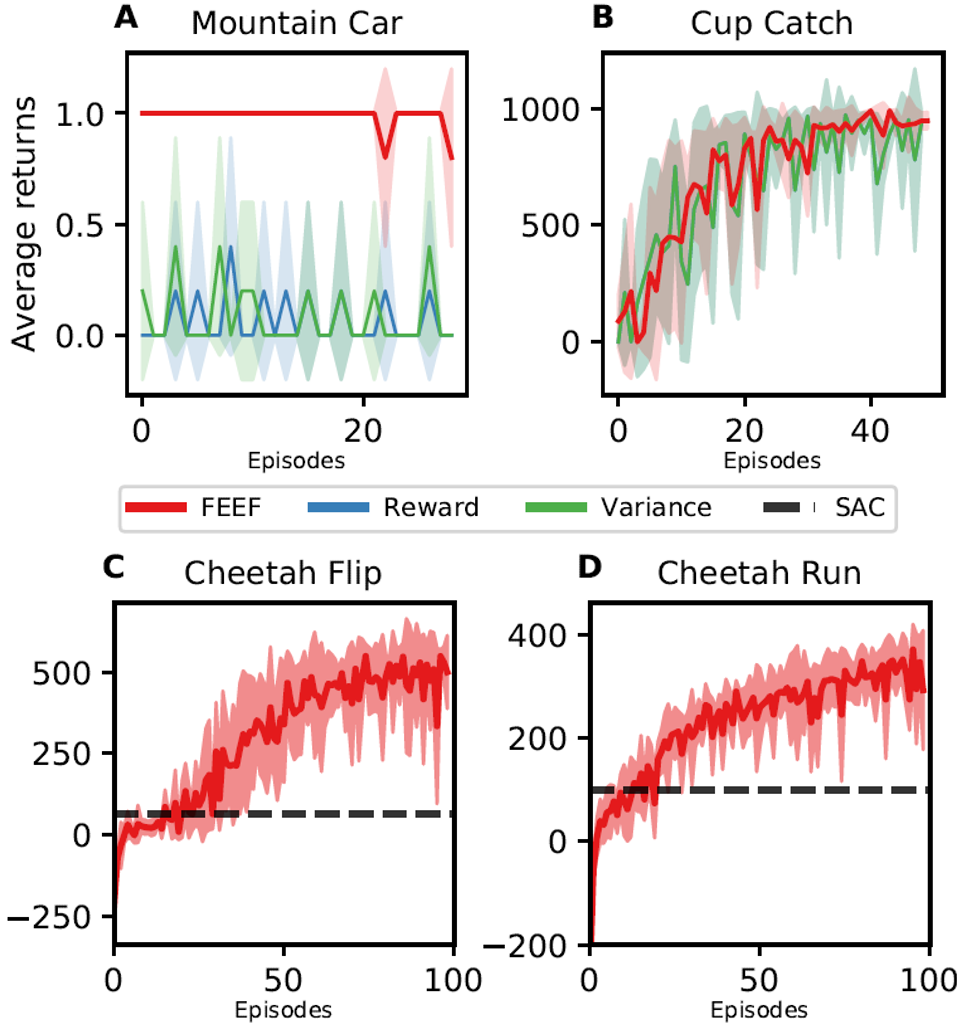}
    \caption{\textbf{AIF planning with rewards}. (A) Mountain Car: Average return after each episode on the sparse-reward Mountain Car task. (B) Cup Catch: Average return after each episode on the sparse-reward Cup Catch task. (C \& D) Half Cheetah: Average return after each episode on the well-shaped Half Cheetah environment, for the running and flipping tasks, respectively. Deep AIF is compared to the average performance of SAC after 100 episodes learning. Each line is the mean of 5 seeds.} \label{fig:planning}
\end{figure}

In \cite{tschantz2020reinforcement}, the authors investigated the performance of deep AIF in the context of \textbf{planning with rewards} in standard MDP RL benchmarks---See Fig. \ref{fig:planning}. Implementation-wise, the deep AIF agent optimizes $q(\pi)$ at each time step, and executes the first action specified by the most likely plan. This involves estimating the EFE for each plan, which in turn involves evaluating the expected future beliefs and observations, given the plan ($q(\mathbf{x}_{k:T}, \mathbf{y}_{k:T}| \pi)$). This can be achieved through the generative model, whereby beliefs about future hidden states (given the current hidden state and plan) are evaluated via the transition distribution, and beliefs about observations, given some hidden state, are evaluated using the likelihood distribution. Given this counterfactual distribution over future states and observations, EFE can be approximated~\cite{millidge2020deep,tschantz2020reinforcement}. The final requirement is an optimization procedure which iteratively updates the plan density such that EFE is minimized. In \cite{tschantz2020reinforcement}, the plan $q(\pi)$ is parameterized as a diagonal Gaussian and the cross-entropy method~\cite{rubinstein1997optimization} is used optimize the parameters such that $q(\pi) \propto -\mathcal{G}(\pi)$. The experiments focused on whether the algorithm was able to \textbf{balance exploration and exploitation}.

The performance was evaluated in domains with (i) well-shaped rewards (\textit{Half Cheetah}), (ii) extremely sparse rewards, where agents only receive reward when the goal is achieved (\textit{Mountain Car} and \textit{Cup Catch}) and (iii) a complete absence of rewards, where there are no rewards and success is measured by the percent of the maze covered (\textit{Ant Maze}). In sparse rewards environment, deep AIF was compared to two baselines, a reward algorithm which only selects plans based on the extrinsic term (ignores the information gain), and a variance algorithm that seeks out uncertain transitions by acting to maximise the output variance of the transition model. For environments with well-shaped rewards, deep AIF was compared to the maximum reward obtained after 100 episodes by a oft-actor-critic (SAC) \cite{haarnoja2018soft}---state-of-the-art model-free RL algorithm.

The Mountain Car experiment is shown in Fig.~\ref{fig:planning}, where we plot the total reward obtained for each episode over 25 episodes, where each episode is at most 200 time steps. These results showed that deep AIF rapidly explores and consistently reaches the goal, achieving optimal performance in a single trial. In contrast, the benchmark algorithms were, on average, unable to successfully explore and achieve good performance. Deep AIF performs comparably to benchmarks on the Cup Catch environment (Fig. 1B). Figure 1 C\&D shows that deep AIF performs substantially better than a state-of-the-art model-free algorithm after 100 episodes on the challenging Half Cheetah tasks.This reflects robust performance in environments with well-shaped rewards and provides considerable improvements in sample-efficiency. The directed exploration afforded by minimizing the EFE proves beneficial in \textbf{environments with no reward structure}. Deep AIF rate of exploration was substantially higher than that of a random baseline in the ant-maze environment, resulting in a more substantial portion of the maze being covered.

New exciting developments and studies of Deep AIF agents are being under research for more complex environments with partial observability and high-dimensional inputs and actions~\cite{fountas2020deep,van_der_Himst_2020,van2021deep,mazzaglia2021contrastive,noel2021online}.

\subsection{Complex cognition}
\label{sec:sota:complex}


\subsubsection{Intention-blended Human-robot collaboration}
\begin{figure}[!hbtp]
    \centering
    \includegraphics[width=0.95\columnwidth, height=310px]{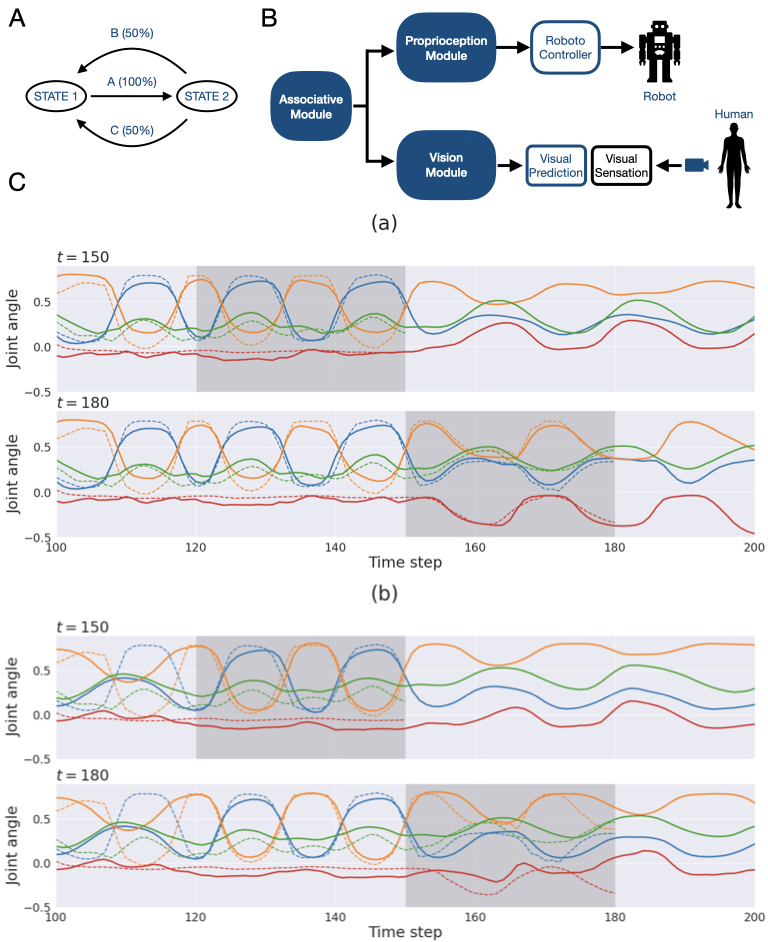}
    \caption{\textbf{Intention-blended Human-robot collaboration}, adapted from \cite{Ohata_2020}. (A) The transition rule of the motion primitives during the imitative interaction. (B) The network model. (C) An example of time-series plots on neural activities in the output layer of the proprioception module in the accuracy term dominant condition (a) and the complexity term dominant condition (b). Reconstruction of the observation and the future prediction at time step 150 (top) and at time step 180 (bottom) are shown in each. Solid and dashed lines represent prediction and observation, respectively. The shadowed area indicates the time window in which the approximate posterior is optimized.} 
    \label{IV-D_figure}
\end{figure}

Ohata and Tani~\cite{Ohata_2020} applied the frameworks of predictive coding and active inference to the social cognition study in which they investigated the dynamics of intention in human-robot interactions. 
In this study, multimodal imitative interactions between a humanoid robot and a human counterpart were simulated in which the robot and human imitate each other's body movement simultaneously. The imitation task was designed to reveal the difference between strong intention and weak intention in social interaction. During imitative interactions sometimes there were cases where the robot tries to perform a different movement primitive than the human. In such conflicting situations, it can be assumed that if the robot has a strong intention, the robot would keep performing the original primitive, and if it has a weak intention, it would change its intention to adapt to the human's primitive. Body movements were implemented using three types of motion primitives (A, B, and C) and they follow specific probabilistic state transition rules. Every time the primitive A comes, either of the primitive B or C follows with 50\% chance, and the primitive A always follows the primitive B and C (Fig. \ref{IV-D_figure}A). Therefore, 

To model multimodal perception and action generation, a hierarchically-organized variational RNN (see Sec.~\ref{sec:aif_learning_hier})
was extended. This model is comprised of three modules: the proprioception module, the vision module, and the associative module (Figure \ref{IV-D_figure}B). 


In the simulation of mutual imitative interaction, the balance between the accuracy and complexity term in the free energy was the key focus. This balance determines how the approximate posterior is optimized through the free energy minimization given the observation and the prior. Results showed that when the complexity term was less dominant, the robot tended to change its prediction, namely intention, easily so that it could adapt to the observation. In a conflicting situation of motion primitives, the robot tended to change its motion primitive to the one the human presented. When the complexity term was more dominant, the robot tended to ignore the observation and retained its original intention. Figure \ref{IV-D_figure}C describes the joint dynamics depending on the complexity term domination. The conflicting situation in which the network predicted the primitive B, but observed C. In the less dominant condition (a), the network reconstructed the observation and modified the prediction such that the primitive C persisted. In the more dominant condition (b), the network ignored the observation and maintained its original prediction.

The congruence between an agent's intention, the action and its anticipated outcome can be linked to the psychology concept of {\it agency}~\cite{gallagher2000philosophical,synofzik2008beyond}. In the context of AIF, an agent's intention can be formulated as a predictive model, and the congruence between the predicted action outcomes and observation could be considered to contribute to the system being in control of the actions and the consequences~\cite{friston2012prediction}. Hence, when the accuracy term dominant condition, the robot is endowed with weaker agency, and in the complexity term dominant condition, the robot owns stronger agency.

\subsubsection{Self/other distinction}
The work in \cite{lanillos2020robot} presented an algorithm that enables a robot to perform non-appearance self/other distinction on a mirror by distinguishing its simple actions from other agents. Developing this visual-kinesthetic matching is essential for safe human-robot interaction, especially in social robotics~\cite{nagai2019predictive,hoffmann2021robot}. Using movement cues the robot first learns the visual forward kinematics and then exploits the Bayesian model evidence to accumulate evidence. The potential of modelling the high-order cognition needed to pass the mirror test in robots makes AIF very attractive for the cognitive sciences view of robotics---See~\cite{hoffmann2021robot} for a computational modelling roadmap.



\section{Connections with other frameworks}
\label{sec:connect}
As one might have realised from previous sections, AIF shares many similarities with other more established control schemes and theories. In this section, we summarise the main commonalities between active inference and other approaches. 
\subsection{Relationship with classical controllers}				\label{sec:connect:control}

Consider the generative model specified by Eq.~\eqref{eq:flaplace}. This model includes the function $f(\latent)$ which determines how the belief state evolves over time. This can be, for example, according to a first order linear system which results in $f(\latent) = (\desiredlatent - \latent)\tau^{-1}$ \cite{buckley2017free,pezzato2020novel,oliver2021empirical}. The belief state is specified to evolve (linearly) over time as the derivative between the current belief $\latent$ and target $\desiredlatent$. The term $\desiredlatent$ indicates the desired state to be achieved while $\tau$ is the time constant. The smaller $\tau$, the larger the derivative. If $\tau$ approaches zero ($\tau^{-1} \rightarrow \infty$), the value $f(\latent)$ approaches $\infty$. As a result, the belief is infinitely biased towards the target and $\latent\approx\desiredlatent$.
A classic PID controller defines an error term $\epsilon = (\desiredlatent - \mathbf{\obs})$. The control action is then chosen as: 
$$a = K_p \epsilon + K_i\!\!\int{\!\epsilon \;dt} +K_d\frac{d\epsilon} {dt}$$
where $K_p, K_i$ and $K_d$ are tuning parameters. For the control law defined by active inference, our $(\mathbf{\obs} - \latent)$ is similar to the error term. Additionally, as explained in the previous section, when $\tau^{-1} \rightarrow  \infty$ then $\latent\approx\desiredlatent$. Now the control law of active inference can be rewritten in terms of the error term as:
$$ \dot{a} = \kappa_{\action} \Sigma_{\obs}^{-1} \epsilon  + \kappa_{\action} \Sigma_{\obs'}^{-1} \frac{d\epsilon}{dt}$$

This means than if $\tau^{-1} \rightarrow \infty$, the active inference controller is equivalent to a PI Controller i.e. PID with $K_d=0$, a $K_p$ gain of $\kappa_{\action} \Sigma_{\obs'}^{-1}$ and an $K_i$ gain of $\kappa_{\action} \Sigma_{\obs}^{-1}$. If one considers the generalized motions up to a third order, the control law would include a non-zero $D$ term. A detailed analysis of how PID arises in active inference under approximate linear generative models can be found in \cite{baltieri2019pid,baioumy2020active}.

Regarding more complex controllers, AIF shares similarities with Linear Quadratic Gaussian control (LQG), since both are grounded in Bayesian inference and optimal control \cite{friston2011optimal}. However, a closer look reveals several key differences between the two approaches regarding the formulation of the state space, the cost functions, and their minimization---See~\cite{baltieri2020kalman}.

AIF can also be considered to be a form of model predictive control~\cite{2020ICRA_baioumy} due to the evaluation of the expected free energy. The key difficulty is evaluating the expectation over all possible future trajectories, which can be approximated via Monte-Carlo sampling of these trajectories using a forward model of the environmental dynamics. Many classical model-predictive control planning algorithms, such as the Cross-Entropy-Method (CEM) \cite{rubinstein1999cross}, and Path-Integral-Control (PI) \cite{theodorou2010generalized,williams2017model} can be used to estimate the value of this integral in the cases where explicitly enumerating every possible future trajectory, as is commonly done in discrete-state-space AIF \cite{friston2017active}. Moreover, it has recently been demonstrated that many of these classical planning algorithms can be interpreted as performing variational inference \cite{okada2020variational}, thus linking model-predictive control closely with the active inference interpretation of adaptive and intelligent behaviour as being fundamentally an inference process. 

A final and very relevant relation can be also found in control as inference~\cite{botvinick2012planning,levine2018reinforcement}---See \cite{millidge2020relationship,watson2020active}. For instance, performing stochastic optimal control when the goal is to infer the input~\cite{watson2020stochastic} can be conceptually regarded as a type of AIF approach.

\subsection{Relationship to Reinforcement Learning}
\label{sec:connect:rl}
Planning with AIF using the expected free energy functional has close relationships to the field of RL. This fact is not unduly surprising since both attempt to solve the same problem of optimal plan selection in unknown environments through largely similar approaches. The key difference between RL and active inference relies on the properties of the EFE. Crucially, the intrinsic and posterior divergence terms of the EFE are not present in RL, only the extrinsic value term is. These differences arise from the fact that active inference is posed as a variational inference procedure, thus requiring a variational approximate distribution, while RL is solely an optimization problem based around maximizing expected reward. Considering only the extrinsic value term, this becomes,
\begin{align}
    \mathbb{E}_{q(\latent_k, \obs_k | \pi)}[\ln \tilde{p}(\obs_k)] \nonumber
\end{align}
Defining the desired distribution to be a Boltzmann distribution around the reward $\tilde{p}(\obs_k) \propto \exp(-r(\obs_k))$, the extrinsic value term simply reduces to the average reward expected under the variational belief distribution in the future. The only remaining difference to RL is that in model-free RL the reward is instead evaluated under the \emph{true} environmental distribution $p(\latent_k, \obs_k)$ instead of under the model distribution, while in model-based RL in practice the reward is evaluated under the model distribution so the equivalence is more exact.

Importantly, however, active inference generalizes and extends reinforcement learning in several important ways. Firstly, the expected free energy objective generalizes the notion of utility by including an intrinsic informational objective, which encourages exploration and equips agents with an `intrinsic curiosity'. It can be shown that this exploratory drive enables agents to perform better in many environments which are high dimensional with sparse rewards, where exploration is necessary to solve the task \cite{tschantz2020reinforcement}. From a mathematical perspective, it is also possible to generalize the expected free energy to a class of objectives called divergence functionals, which all result in this combination of reward-seeking and information-seeking behaviour \cite{millidge2021whence,millidge2021understanding}. Secondly, active inference's notion of encoding goals as a prior distribution over observations is more flexible than the use of rewards in reinforcement learning, since, as shown above, RL implicitly assumes that the prior distribution is Boltzmann, while active inference is free to use any other prior distribution instead which enables considerably more flexibility in specifying goals.


\section{Benefits and Challenges}
\label{sec:conclusion}

For robot perception, control and learning, Active Inference is: 1) a unified framework, 2) with functional biological plausibility, 3) using variational Bayesian inference. Each of these three aspects lead to specific expected benefits for robot perception and control, as well as to interesting open research challenges.

\subsection{Unified framework}
\label{sec:conclusion:unified}
Perhaps the most exciting aspect of AIF is the natural integration of perception and action into a single objective, namely the minimization of Variational Free Energy (or the Expected Free Energy when planning). A potential result is that an AIF agent could use action to make its world behave more predictably. This introduces, in a sense, a double model-bias; not only is state estimation recursively biased towards a Bayesian prior, even the actions help reinforce this bias. In other words, rather than trying to accurately model the world, and requiring unfathomable amounts of data to do so, the AIF agent gets by with a strongly simplified model which it enforces (where possible) through its own actions. This may hold an essential key to solve the data- and experience hungriness of present-day state-of-the-art learning algorithms. 

There are several interesting research challenges still to be solved, related to this model-bias. First, the question is how to avoid suboptimal convergence \cite{deisenroth2011pilco}, possibly through the presence of intrinsic value in the optimization criterion \cite{tschantz2020learning}. Second, if the action consists of some continuous-time feedback signal, it is a challenge to find the update rule for (the parameters of) that feedback signal. For simple systems, the observation w.r.t. action derivative of the VFE can be directly computed (e.g. \cite{buckley2017free}). However, this is very challenging for complex systems---see the closed form equation in \cite[appendix]{friston2010action} when the system is known---and hence it is usually approximated with a directional constant~\cite{oliver2021empirical,baioumy2020active, pezzato2020active} or by performing the control on the proprioceptive states~\cite{oliver2021empirical,meo2021multimodal}. Potentially interesting alternative solutions exist in the concepts of adaptive interaction~\cite{lin2000self} or direct gradient descent control~\cite{naiborhu2006direct}. Finally, there is a challenging question regarding the purpose of the belief (internal state) $\latent$, which in vanilla AIF is biased towards the desired state. This means, unless the agent is at the target state, the belief will be biased, and therefore does not accurately represent the actual hidden state.
A potential solution might consist of introducing the action in the generative model~\cite{watson2020stochastic,baioumy2021fault} in the same way that we included it in the planning. 

As a second expected benefit, AIF not only integrates action with perception, it also provides for a natural Bayesian integration of intrinsic and extrinsic value into a single planning objective. This is formally appealing. An interesting remaining research challenge is to find a principled way to balance intrinsic and extrinsic value, regulated in AIF by the precision of the priors on desired observations~\cite{van2021deep}, to completely remove the need for a heuristic exploration/exploitation trade-off.

Finally, AIF provides a way to integrate generative models in a hierarchical fashion aiding the construction of complex probabilistic controllers~\cite{friston2017graphical}.

\subsection{Functional biological plausibility}

The concept of AIF has a very strong presence in the field of neuroscience with an ever-increasing list of showcases of functional biological plausibility of the concept~\cite{friston2017active}. This naturally leads to great expectations for the field of robotics; if AIF is indeed an accurate mathematical description of the neurological processes underlying biological perception, control, and most importantly cognition, then it may bring similar cognitive skills to our robots. Specifically, the AIF approach (and the general framework of predictive coding) allows, by construction~\cite{sandved2021towards}, to go beyond control and achieve high-order cognitive and metacognitive capabilities, such as monitoring, self-explainability and in some degree ``awareness".

In a broader sense of metacognition, i.e., cognition about cognition~\cite{Cleeremans-TCS2019}, the robot should be able to evaluate and monitor the first-order cognitive processes~\cite{hesp2021deeply}. One distinct characteristics of the AIF is that it uses the second-order formula in terms of the precision in prediction. The predictive model does not just predict sensations but also their predictability. This means that robots can become self-attentive and monitor its uncertainty. This is relevant for applications, such as human-robot interaction in industrial~\cite{Ohata_2020} and healthcare settings. 

Furthermore, there are interesting connections between AIF and two awareness abilities in humans that may be essential for interaction, and would bring to robotics a way to enforce explainability and safety. First, self-awareness, where the agent is able to differentiate as an independent entity. or instance, Tani and colleagues~\cite{tani1998interpretation,tani2020cognitive} as well as Lanillos et al.~\cite{lanillosRobotSelfOther2020,hoffmann2021robot} proposed that the prediction/reconstruction error for the past sensation could be related to the sense of self-awareness in machines. Second, agency, i.e., the feeling of controlling the actions and consequences. Hence, robotic solutions for body self-awareness~\cite{lanillos2021neuroscience,hinz2018drifting}, and models of agency are exciting opportunities under the AIF approach.


Additionally, there is an open research question of a very different nature, but also connected to the biological plausibility of Active Inference. Inherited from Friston's world-renowned work on dynamic causal modeling (DCM) for brain-imaging~\cite{friston2003dynamic}, AIF uses the concept of "generalized coordinates", i.e. appending system states with up to the 6$^{th}$ order of derivatives of those states, which is quite alien for control engineers. We have shown that this improves estimation accuracy in drones in wind~\cite{ajith2021DEM_drone}, for example. The open research question is to what extend the use of generalized coordinates is worth the methodological investment for generic robotic applications, and whether it can be replaced by extending system model states with noise filter model states.

\subsection{Variational Bayesian Inference}
The fundamental mathematical operation of AIF is variational Bayesian Inference, an approach which has already been gaining popularity within the robotics community~\cite{toussaint2009robot,deisenroth2013survey,levine2018reinforcement}. The expected benefit is that, when a proper (hierarchical) set of Bayesian priors is in place, robots will be able to perceive, decide, and learn with much less data or require much fewer trials that current systems. To obtain such a proper set of priors, we expect that AIF will provide a highly natural framework for humans to teach robots; the required hierarchical set of Bayesian priors can possibly be obtained through a proper curriculum of demonstration and training. This is still a wide open and very exciting research area.

Second, we expect that the variational Bayesian inference approach will lead to a leap forward in \textit{fault monitoring and fault tolerance}. If the entire control hierarchy is based on prediction errors, then all unexpected sensory inputs will be noticed. By maintaining not only state-dependent expectations of the sensor value itself, but also expectations of the variance of those values, the system will only trigger on signals that are outside regular bounds. As a follow-up from our preliminary robot arm experiments, we expect eventually to see robots that predict all sensor signals at all times, and detect any unexpected behaviour, including (unexpected) collisions, sensor/actuator malfunction, sample frequency hiccups, and even unexpected user behavior.

Third, we expect that the variational Bayesian inference approach will help alleviate the combinatorial explosion associated with making longer-term plans, and the accompanying deterioration in accuracy of predictions with the number planning steps. In principle this should be an emergent property of AIF, in the sense the the objective to minimize free energy entails a minimization of complexity, both statistically and in terms of computational cost via the Jarzynski equality (see~\cite{friston2020sophisticated} for an approach along these lines). AIF suggests possible approaches to address both issues. For example, it suggests a principled solution to combinatorial explosion of  plans in terms of an approximate  factorisation of the  distributions over actions, a technique core to variational approaches in general. Uncertainty accumulated during the execution of a given policy can be quantified and thus actions planned accordingly. The generality of the variational Bayesian formulation provides an intuitive foundation from which to develop new uncertainty-sensitive approaches and gives a very appealing probabilistic solution to model-predictive control.

\section{Summary}
We discussed how a theory of cognition originating in computational neuroscience opened up opportunities to improve robotic systems. In particular, we detailed its application in estimation, adaptive control, planning and learning. We described both the mathematical formulation of AIF and the most relevant works in the literature as well as showcasing some experiments and lessons learned from deploying  AIF in real robotic systems, such as industrial manipulators or humanoids. We also described its connection with other fields like classical control and RL. Finally, we discussed the benefits and challenges of this approach to transform AIF into a standard methodology in robotics and to give robots human-like interactive capabilities.

\section*{Acknowledgment}
We would like to thank Karl Friston for his comments on the manuscript and his invaluable inspiration for AIF in robotics.

\ifCLASSOPTIONcaptionsoff
  \newpage
\fi


\bibliographystyle{IEEEtran}
\bibliography{references,pl}

\begin{thebibliography}{100}
\providecommand{\url}[1]{#1}
\csname url@samestyle\endcsname
\providecommand{\newblock}{\relax}
\providecommand{\bibinfo}[2]{#2}
\providecommand{\BIBentrySTDinterwordspacing}{\spaceskip=0pt\relax}
\providecommand{\BIBentryALTinterwordstretchfactor}{4}
\providecommand{\BIBentryALTinterwordspacing}{\spaceskip=\fontdimen2\font plus
\BIBentryALTinterwordstretchfactor\fontdimen3\font minus
  \fontdimen4\font\relax}
\providecommand{\BIBforeignlanguage}[2]{{%
\expandafter\ifx\csname l@#1\endcsname\relax
\typeout{** WARNING: IEEEtran.bst: No hyphenation pattern has been}%
\typeout{** loaded for the language `#1'. Using the pattern for}%
\typeout{** the default language instead.}%
\else
\language=\csname l@#1\endcsname
\fi
#2}}
\providecommand{\BIBdecl}{\relax}
\BIBdecl

\bibitem{friston2010free}
K.~Friston, ``The free-energy principle: a unified brain theory?'' \emph{Nature
  reviews neuroscience}, vol.~11, no.~2, pp. 127--138, 2010.

\bibitem{Helmholtz1867}
H.~v. Helmholtz, \emph{Handbuch der physiologischen Optik}.\hskip 1em plus
  0.5em minus 0.4em\relax L. Voss, 1867.

\bibitem{kmd1951puzzle}
K.~M. Dallenbach, ``A puzzle-picture with a new principle of concealment,''
  \emph{The American journal of psychology}, pp. 431--433, 1951.

\bibitem{rao1999predictive}
R.~P. Rao and D.~H. Ballard, ``Predictive coding in the visual cortex: a
  functional interpretation of some extra-classical receptive-field effects,''
  \emph{Nature neuroscience}, vol.~2, no.~1, pp. 79--87, 1999.

\bibitem{buckley2017free}
C.~L. Buckley, C.~S. Kim, S.~McGregor, and A.~K. Seth, ``The free energy
  principle for action and perception: A mathematical review,'' \emph{Journal
  of Mathematical Psychology}, vol.~81, pp. 55--79, 2017.

\bibitem{lanillos2021neuroscience}
P.~Lanillos and M.~van Gerven, ``Neuroscience-inspired perception-action in
  robotics: applying active inference for state estimation, control and
  self-perception,'' \emph{arXiv preprint arXiv:2105.04261}, 2021.

\bibitem{lanillos2020predictive}
P.~Lanillos, S.~Franklin, A.~Maselli, and D.~W. Franklin, ``Active strategies
  for multisensory conflict suppression in the virtual hand illusion,''
  \emph{Scientific Reports}, 2021.

\bibitem{e23101306}
A.~Anil~Meera and M.~Wisse, ``Dynamic expectation maximization algorithm for
  estimation of linear systems with colored noise,'' \emph{Entropy}, vol.~23,
  no.~10, 2021.

\bibitem{meera2020free}
A.~A. Meera and M.~Wisse, ``Free energy principle based state and input
  observer design for linear systems with colored noise,'' in \emph{2020
  American Control Conference (ACC)}.\hskip 1em plus 0.5em minus 0.4em\relax
  IEEE, 2020, pp. 5052--5058.

\bibitem{ajith2021DEM_drone}
A.~Anil~Meera and M.~Wisse, ``A brain inspired learning algorithm for the
  perception of a quadrotor in wind,'' \emph{arXiv preprint arXiv:2109.11971},
  2021.

\bibitem{bos2021free}
F.~Bos, A.~Anil~Meera, D.~Benders, and M.~Wisse, ``Free energy principle for
  state and input estimation of a quadcopter flying in wind,'' \emph{arXiv
  preprint arXiv:2109.12052}, 2021.

\bibitem{lanillos2018adaptive}
P.~Lanillos and G.~Cheng, ``Adaptive robot body learning and estimation through
  predictive coding,'' in \emph{2018 IEEE/RSJ International Conference on
  Intelligent Robots and Systems (IROS)}.\hskip 1em plus 0.5em minus
  0.4em\relax IEEE, 2018, pp. 4083--4090.

\bibitem{oliver2021empirical}
G.~Oliver, P.~Lanillos, and G.~Cheng, ``An empirical study of active inference
  on a humanoid robot,'' \emph{IEEE Transactions on Cognitive and Developmental
  Systems}, 2021.

\bibitem{burghardt2021robot}
D.~Burghardt and P.~Lanillos, ``Robot localization and navigation through
  predictive processing using lidar,'' \emph{arXiv preprint arXiv:2109.04139},
  2021.

\bibitem{pio2016active}
L.~Pio-Lopez, A.~Nizard, K.~Friston, and G.~Pezzulo, ``Active inference and
  robot control: a case study,'' \emph{Journal of The Royal Society Interface},
  vol.~13, no. 122, p. 20160616, 2016.

\bibitem{Lanillos2018ActiveIW}
P.~Lanillos and G.~Cheng, ``Active inference with function learning for robot
  body perception,'' in \emph{Proc. Int. Workshop Continual Unsupervised
  Sensorimotor Learn.}, 2018, pp. 1--5.

\bibitem{pezzato2020novel}
C.~Pezzato, R.~Ferrari, and C.~H. Corbato, ``A novel adaptive controller for
  robot manipulators based on active inference,'' \emph{IEEE Robotics and
  Automation Letters}, vol.~5, no.~2, pp. 2973--2980, 2020.

\bibitem{baioumy2020active}
M.~Baioumy, P.~Duckworth, B.~Lacerda, and N.~Hawes, ``Active inference for
  integrated state-estimation, control, and learning,'' in \emph{International
  conference on Robotics and Automation, ICRA}, 2021.

\bibitem{sancaktar2020end}
C.~Sancaktar, M.~van Gerven, and P.~Lanillos, ``End-to-end pixel-based deep
  active inference for body perception and action,'' in \emph{2020 Joint IEEE
  10th International Conference on Development and Learning and Epigenetic
  Robotics (ICDL-EpiRob)}, 2020.

\bibitem{meo2021multimodal}
C.~Meo and P.~Lanillos, ``Multimodal vae active inference controller,'' in
  \emph{2021 IEEE/RSJ International Conference on Intelligent Robots and
  Systems (IROS)}.\hskip 1em plus 0.5em minus 0.4em\relax IEEE, 2021.

\bibitem{rood2020deep}
T.~Rood, M.~van Gerven, and P.~Lanillos, ``A deep active inference model of the
  rubber-hand illusion,'' \emph{arXiv preprint arXiv:2008.07408}, 2020.

\bibitem{baioumy2021fault}
M.~Baioumy, C.~Pezzato, R.~Ferrari, C.~H. Corbato, and N.~Hawes,
  ``Fault-tolerant control of robot manipulators with sensory faults using
  unbiased active inference,'' in \emph{European Control Conference, ECC},
  2021.

\bibitem{pezzato2020active}
C.~Pezzato, M.~Baioumy, C.~H. Corbato, N.~Hawes, M.~Wisse, and R.~Ferrari,
  ``Active inference for fault tolerant control of robot manipulators with
  sensory faults,'' in \emph{International Workshop on Active Inference}.\hskip
  1em plus 0.5em minus 0.4em\relax Springer, 2020, pp. 20--27.

\bibitem{Baltieri2017Active}
\emph{{An active inference implementation of phototaxis}}, ser. ALIFE 2021: The
  2021 Conference on Artificial Life, vol. ECAL 2017, the Fourteenth European
  Conference on Artificial Life, 09 2017.

\bibitem{friston2009reinforcement}
K.~J. Friston, J.~Daunizeau, and S.~J. Kiebel, ``Reinforcement learning or
  active inference?'' \emph{PloS one}, vol.~4, no.~7, p. e6421, 2009.

\bibitem{millidge2020reinforcement}
B.~Millidge, A.~Tschantz, A.~K. Seth, and C.~L. Buckley, ``Reinforcement
  learning as iterative and amortised inference,'' 2020.

\bibitem{sajid2019active}
N.~Sajid, P.~J. Ball, and K.~J. Friston, ``Active inference: demystified and
  compared,'' \emph{arXiv}, pp. arXiv--1909, 2019.

\bibitem{tschantz2020scaling}
A.~Tschantz, M.~Baltieri, A.~K. Seth, and C.~L. Buckley, ``Scaling active
  inference,'' in \emph{2020 International Joint Conference on Neural Networks
  (IJCNN)}.\hskip 1em plus 0.5em minus 0.4em\relax IEEE, 2020, pp. 1--8.

\bibitem{Han-Tani-RL-AIF-2021}
D.~Han, K.~Doya, and J.~Tani, ``Goal-directed planning by reinforcement
  learning and active inference,'' \emph{arXiv preprint arxiv:2106.09938v2},
  2021.

\bibitem{millidge2019combining}
B.~Millidge, ``Combining active inference and hierarchical predictive coding: A
  tutorial introduction and case study,'' 2019.

\bibitem{tschantz2020learning}
A.~Tschantz, A.~K. Seth, and C.~L. Buckley, ``Learning action-oriented models
  through active inference,'' \emph{PLoS computational biology}, vol.~16,
  no.~4, p. e1007805, 2020.

\bibitem{friston2020sophisticated}
K.~Friston, L.~Da~Costa, D.~Hafner, C.~Hesp, and T.~Parr, ``Sophisticated
  inference,'' \emph{arXiv preprint arXiv:2006.04120}, 2020.

\bibitem{ueltzhoffer2018deep}
K.~Ueltzh{\"o}ffer, ``Deep active inference,'' \emph{Biological cybernetics},
  vol. 112, no.~6, pp. 547--573, 2018.

\bibitem{van_der_Himst_2020}
O.~van~der Himst and P.~Lanillos, ``Deep active inference for partially
  observable mdps,'' \emph{Communications in Computer and Information Science},
  p. 61–71, 2020.

\bibitem{fountas2020deep}
Z.~Fountas, N.~Sajid, P.~A.~M. Mediano, and K.~Friston, ``Deep active inference
  agents using monte-carlo methods,'' 2020.

\bibitem{millidge2020deep}
B.~Millidge, ``Deep active inference as variational policy gradients,''
  \emph{Journal of Mathematical Psychology}, vol.~96, p. 102348, 2020.

\bibitem{ccatal2020learning}
O.~{\c{C}}atal, S.~Wauthier, C.~De~Boom, T.~Verbelen, and B.~Dhoedt, ``Learning
  generative state space models for active inference,'' \emph{Frontiers in
  Computational Neuroscience}, vol.~14, p. 103, 2020.

\bibitem{catal2020deep}
O.~{\c{C}}atal, S.~Wauthier, T.~Verbelen, C.~De~Boom, and B.~Dhoedt, ``Deep
  active inference for autonomous robot navigation,'' in \emph{The bridging AI
  and cognitive science (BAICS) workshop, ICLR}, 2020.

\bibitem{catal2021robot}
O.~{\c{C}}atal, T.~Verbelen, T.~Van~de Maele, B.~Dhoedt, and A.~Safron, ``Robot
  navigation as hierarchical active inference,'' \emph{Neural Networks}, vol.
  142, pp. 192--204, 2021.

\bibitem{Matsumoto_2020}
T.~Matsumoto and J.~Tani, ``Goal-directed planning for habituated agents by
  active inference using a variational recurrent neural network,''
  \emph{Entropy}, vol.~22, no.~5, p. 564, May 2020.

\bibitem{Traub2021Dynamic}
M.~Traub, M.~V. Butz, R.~Legenstein, and S.~Otte, ``Dynamic action inference
  with recurrent spiking neural networks,'' in \emph{Artificial Neural Networks
  and Machine Learning -- ICANN 2021}, I.~Farka{\v{s}}, P.~Masulli, S.~Otte,
  and S.~Wermter, Eds.\hskip 1em plus 0.5em minus 0.4em\relax Cham: Springer
  International Publishing, 2021, pp. 233--244.

\bibitem{pezzato2020activeBT}
C.~Pezzato, C.~Hernandez, S.~Bonhof, and M.~Wisse, ``Active inference and
  behavior trees for reactive action planning and execution in robotics,''
  \emph{arXiv preprint arXiv:2011.09756}, 2020.

\bibitem{murata-tani2015}
S.~Murata, Y.~Yamashita, H.~Arie, T.~Ogata, S.~Sugano, and J.~Tani, ``Learning
  to perceive the world as probabilistic or deterministic via interaction with
  others: A neuro-robotics experiment,'' \emph{IEEE transactions on neural
  networks and learning systems}, vol.~28, no.~4, pp. 830--848, 2015.

\bibitem{Ohata_2020}
W.~Ohata and J.~Tani, ``Investigation of the sense of agency in social
  cognition, based on frameworks of predictive coding and active inference: A
  simulation study on multimodal imitative interaction,'' \emph{Frontiers in
  Neurorobotics}, vol.~14, Sep 2020.

\bibitem{chame2020AHybrid}
H.~F. Chame, A.~Ahmadi, and J.~Tani, ``A hybrid human-neurorobotics approach to
  primary intersubjectivity via active inference,'' \emph{Frontiers in
  Psychology}, vol.~11, p. 3207, 2020.

\bibitem{chame2020cognitive}
H.~F. Chame and J.~Tani, ``Cognitive and motor compliance in intentional
  human-robot interaction,'' 2020.

\bibitem{lanillosRobotSelfOther2020}
P.~Lanillos, J.~Pages, and G.~Cheng, ``Robot self/other distinction: Active
  inference meets neural networks learning in a mirror,'' in \emph{European
  Conference on Artificial Intelligence}.\hskip 1em plus 0.5em minus
  0.4em\relax Amsterdam: IOS Press, 2020.

\bibitem{hoffmann2021robot}
M.~Hoffmann, S.~Wang, V.~Outrata, E.~Alzueta, and P.~Lanillos, ``Robot in the
  mirror: toward an embodied computational model of mirror self-recognition,''
  \emph{KI-K{\"u}nstliche Intelligenz}, vol.~35, no.~1, pp. 37--51, 2021.

\bibitem{kirchhoff2018markov}
M.~Kirchhoff, T.~Parr, E.~Palacios, K.~Friston, and J.~Kiverstein, ``The markov
  blankets of life: autonomy, active inference and the free energy principle,''
  \emph{Journal of The royal society interface}, vol.~15, no. 138, p. 20170792,
  2018.

\bibitem{bogacz2017tutorial}
R.~Bogacz, ``A tutorial on the free-energy framework for modelling perception
  and learning,'' \emph{Journal of mathematical psychology}, vol.~76, pp.
  198--211, 2017.

\bibitem{baltieri2019pid}
M.~Baltieri and C.~L. Buckley, ``Pid control as a process of active inference
  with linear generative models,'' \emph{Entropy}, vol.~21, no.~3, p. 257,
  2019.

\bibitem{vandelaar2020application}
T.~van~de Laar, A.~Özçelikkale, and H.~Wymeersch, ``Application of the free
  energy principle to estimation and control,'' 2020.

\bibitem{2020ICRA_baioumy}
M.~Baioumy, M.~Mattamala, and N.~Hawes, ``Variational inference for predictive
  and reactive controllers,'' in \emph{ICRA 2020 Workshop on New advances in
  Brain-inspired Perception, Interaction and Learning}, Paris, France, 2020.

\bibitem{da2020active}
L.~Da~Costa, T.~Parr, N.~Sajid, S.~Veselic, V.~Neacsu, and K.~Friston, ``Active
  inference on discrete state-spaces: a synthesis,'' \emph{Journal of
  Mathematical Psychology}, vol.~99, p. 102447, 2020.

\bibitem{tschantz2020reinforcement}
A.~Tschantz, B.~Millidge, A.~K. Seth, and C.~L. Buckley, ``Reinforcement
  learning through active inference,'' \emph{arXiv preprint arXiv:2002.12636},
  2020.

\bibitem{Friston2008DEM}
K.~Friston, N.~Trujillo-Barreto, and J.~Daunizeau, ``Dem: A variational
  treatment of dynamic systems,'' \emph{NeuroImage}, vol.~41, pp. 849--85, 08
  2008.

\bibitem{ciria2021predictive}
A.~Ciria, G.~Schillaci, G.~Pezzulo, V.~V. Hafner, and B.~Lara, ``Predictive
  processing in cognitive robotics: a review,'' 2021.

\bibitem{Spratling2017review}
M.~Spratling, ``A review of predictive coding algorithms,'' \emph{Brain and
  Cognition}, vol. 112, pp. 92--97, 2017, perspectives on Human Probabilistic
  Inferences and the 'Bayesian Brain'.

\bibitem{PEZZULO201517}
G.~Pezzulo, F.~Rigoli, and K.~Friston, ``Active inference, homeostatic
  regulation and adaptive behavioural control,'' \emph{Progress in
  Neurobiology}, vol. 134, pp. 17--35, 2015.

\bibitem{jordan1999introduction}
M.~I. Jordan, Z.~Ghahramani, T.~S. Jaakkola, and L.~K. Saul, ``An introduction
  to variational methods for graphical models,'' \emph{Machine learning},
  vol.~37, no.~2, pp. 183--233, 1999.

\bibitem{Friston2007Variational}
K.~Friston, J.~Mattout, N.~Trujillo-Barreto, J.~Ashburner, and W.~Penny,
  ``Variational free energy and the laplace approximation,'' \emph{NeuroImage},
  vol.~34, pp. 220--34, 02 2007.

\bibitem{buckleyFEP}
C.~Buckley, C.~Kim, S.~McGregor, and A.~Seth, ``The free energy principle for
  action and perception: A mathematical review,'' \emph{Journal of Mathematical
  Psychology}, vol.~81, pp. 55--79, 2017.

\bibitem{friston2010action}
K.~J. Friston, J.~Daunizeau, J.~Kilner, and S.~J. Kiebel, ``Action and
  behavior: a free-energy formulation,'' \emph{Biological cybernetics}, vol.
  102, no.~3, pp. 227--260, 2010.

\bibitem{friston2009predictive}
K.~Friston and S.~Kiebel, ``Predictive coding under the free-energy
  principle,'' \emph{Philosophical transactions of the Royal Society B:
  Biological sciences}, vol. 364, no. 1521, pp. 1211--1221, 2009.

\bibitem{friston2008variational}
K.~J. Friston, N.~Trujillo-Barreto, and J.~Daunizeau, ``Dem: a variational
  treatment of dynamic systems,'' \emph{Neuroimage}, vol.~41, no.~3, pp.
  849--885, 2008.

\bibitem{ajith2021convergence}
A.~Anil~Meera and M.~Wisse, ``On the convergence of dem's linear parameter
  estimator,'' in \emph{International Workshop on Active Inference}, 2021.

\bibitem{kaelbling1998planning}
L.~P. Kaelbling, M.~L. Littman, and A.~R. Cassandra, ``Planning and acting in
  partially observable stochastic domains,'' \emph{Artificial intelligence},
  vol. 101, no. 1-2, pp. 99--134, 1998.

\bibitem{friston2015active}
K.~Friston, F.~Rigoli, D.~Ognibene, C.~Mathys, T.~Fitzgerald, and G.~Pezzulo,
  ``Active inference and epistemic value,'' \emph{Cognitive neuroscience},
  vol.~6, no.~4, pp. 187--214, 2015.

\bibitem{parr2019generalised}
T.~Parr and K.~J. Friston, ``Generalised free energy and active inference,''
  \emph{Biological cybernetics}, vol. 113, no.~5, pp. 495--513, 2019.

\bibitem{millidge2021whence}
B.~Millidge, A.~Tschantz, and C.~L. Buckley, ``Whence the expected free
  energy?'' \emph{Neural Computation}, vol.~33, no.~2, pp. 447--482, 2021.

\bibitem{friston2017active}
K.~Friston, T.~FitzGerald, F.~Rigoli, P.~Schwartenbeck, and G.~Pezzulo,
  ``Active inference: a process theory,'' \emph{Neural computation}, vol.~29,
  no.~1, pp. 1--49, 2017.

\bibitem{schmidhuber2010formal}
J.~Schmidhuber, ``Formal theory of creativity, fun, and intrinsic motivation
  (1990--2010),'' \emph{IEEE Transactions on Autonomous Mental Development},
  vol.~2, no.~3, pp. 230--247, 2010.

\bibitem{oudeyer2009intrinsic}
P.-Y. Oudeyer and F.~Kaplan, ``What is intrinsic motivation? a typology of
  computational approaches,'' \emph{Frontiers in neurorobotics}, vol.~1, p.~6,
  2009.

\bibitem{schwartenbeck2019computational}
P.~Schwartenbeck, J.~Passecker, T.~U. Hauser, T.~H. FitzGerald, M.~Kronbichler,
  and K.~J. Friston, ``Computational mechanisms of curiosity and goal-directed
  exploration,'' \emph{Elife}, vol.~8, p. e41703, 2019.

\bibitem{lanillos2018active}
P.~Lanillos and G.~Cheng, ``Active inference with function learning for robot
  body perception,'' \emph{International Workshop on Continual Unsupervised
  Sensorimotor Learning, IEEE Developmental Learning and Epigenetic Robotics
  (ICDL-Epirob)}, 2018.

\bibitem{parr2021generative}
T.~Parr, N.~Sajid, L.~Da~Costa, M.~B. Mirza, and K.~J. Friston, ``Generative
  models for active vision,'' \emph{Frontiers in Neurorobotics}, vol.~15,
  p.~34, 2021.

\bibitem{sajid2021active}
N.~Sajid, P.~J. Ball, T.~Parr, and K.~J. Friston, ``Active inference:
  demystified and compared,'' \emph{Neural Computation}, vol.~33, no.~3, pp.
  674--712, 2021.

\bibitem{catal2019bayesian}
O.~Çatal, J.~Nauta, T.~Verbelen, P.~Simoens, and B.~Dhoedt, ``Bayesian policy
  selection using active inference,'' 2019.

\bibitem{van2021deep}
N.~van Hoeffelen and P.~Lanillos, ``Deep active inference for pixel-based
  discrete control: Evaluation on the car racing problem,'' \emph{arXiv
  preprint arXiv:2109.04155}, 2021.

\bibitem{yamashita-tani2008}
Y.~Yamashita and J.~Tani, ``Emergence of functional hierarchy in a multiple
  timescale neural network model: a humanoid robot experiment,'' \emph{PLoS
  computational biology}, vol.~4, no.~11, p. e1000220, 2008.

\bibitem{ahmadi-tani2019}
A.~Ahmadi and J.~Tani, ``A novel predictive-coding-inspired variational rnn
  model for online prediction and recognition,'' \emph{Neural computation},
  vol.~31, no.~11, pp. 2025--2074, 2019.

\bibitem{chung2015recurrent}
J.~Chung, K.~Kastner, L.~Dinh, K.~Goel, A.~C. Courville, and Y.~Bengio, ``A
  recurrent latent variable model for sequential data,'' in \emph{Advances in
  neural information processing systems}, 2015, pp. 2980--2988.

\bibitem{rumelhart1985learning}
D.~E. Rumelhart, G.~E. Hinton, and R.~J. Williams, ``Learning internal
  representations by error propagation,'' California Univ San Diego La Jolla
  Inst for Cognitive Science, Tech. Rep., 1985.

\bibitem{chameICRA2020}
H.~F. Chame and J.~Tani, ``Cognitive and motor compliance in intentional
  human-robot interaction,'' in \emph{Proc. of 2020 IEEE International
  Conference on Robotics and Automation (ICRA)}, 2020, pp. 11\,291--11\,297.

\bibitem{chameFrontier2020}
H.~F. Chame, A.~Ahmadi, and J.~Tani, ``A hybrid human-neurorobotics approach to
  primary intersubjectivity via active inference,'' \emph{Frontiers in
  Psychology}, vol.~11, p. 3207, 2020.

\bibitem{wirkuttis-tani2021}
N.~Wirkuttis and J.~Tani, ``Leading or following? dyadic robot imitative
  interaction using the active inference framework,'' \emph{IEEE Robotics and
  Automation Letters}, vol.~6, no.~3, pp. 6024--6031, 2021.

\bibitem{matsumoto2020goal}
T.~Matsumoto and J.~Tani, ``Goal-directed planning for habituated agents by
  active inference using a variational recurrent neural network,''
  \emph{Entropy}, vol.~22, no.~5, p. 564, 2020.

\bibitem{tani2003}
J.~Tani, ``Learning to generate articulated behavior through the bottom-up and
  the top-down interaction processes,'' \emph{Neural networks}, vol.~16, no.~1,
  pp. 11--23, 2003.

\bibitem{tani2004}
J.~Tani, M.~Ito, and Y.~Sugita, ``Self-organization of distributedly
  represented multiple behavior schemata in a mirror system: reviews of robot
  experiments using rnnpb,'' \emph{Neural Networks}, vol.~17, no. 8-9, pp.
  1273--1289, 2004.

\bibitem{friston-biol-2011}
K.~Friston, J.~Mattout, and J.~Kilner, ``Action understanding and active
  inference,'' \emph{Biological cybernetics}, vol. 104, no.~1, pp. 137--160,
  2011.

\bibitem{hinz2018drifting}
N.-A. Hinz, P.~Lanillos, H.~Mueller, and G.~Cheng, ``Drifting perceptual
  patterns suggest prediction errors fusion rather than hypothesis selection:
  replicating the rubber-hand illusion on a robot,'' in \emph{2018 Joint IEEE
  8th International Conference on Development and Learning and Epigenetic
  Robotics (ICDL-EpiRob)}.\hskip 1em plus 0.5em minus 0.4em\relax IEEE, 2018,
  pp. 125--132.

\bibitem{astrom}
K.~Astrom, ``Theory and applications of adaptive control - a survey,''
  \emph{Automatica}, vol. Vol. 19, No. 5, pp. 471--486, 1983.

\bibitem{colledanchise2017}
M.~Colledanchise and P.~Ogren, ``{How Behavior Trees Modularize Hybrid Control
  Systems and Generalize Sequential Behavior Compositions, the Subsumption
  Architecture, and Decision Trees},'' \emph{IEEE Transactions on Robotics},
  vol.~33, no.~2, pp. 372--389, 2017.

\bibitem{baioumyIWAI2021}
M.~Baioumy, C.~Pezzato, C.~H. Corbato, N.~Hawes, and R.~Ferrari, ``Towards
  stochastic fault-tolerant control usingprecision learning and active
  inference,'' in \emph{International Workshop on Active Inference}.\hskip 1em
  plus 0.5em minus 0.4em\relax Springer, 2021.

\bibitem{friston2017graphical}
K.~J. Friston, T.~Parr, and B.~de~Vries, ``The graphical brain: belief
  propagation and active inference,'' \emph{Network Neuroscience}, vol.~1,
  no.~4, pp. 381--414, 2017.

\bibitem{rubinstein1997optimization}
R.~Y. Rubinstein, ``Optimization of computer simulation models with rare
  events,'' \emph{European Journal of Operational Research}, vol.~99, no.~1,
  pp. 89--112, 1997.

\bibitem{haarnoja2018soft}
T.~Haarnoja, A.~Zhou, P.~Abbeel, and S.~Levine, ``Soft actor-critic: Off-policy
  maximum entropy deep reinforcement learning with a stochastic actor,'' in
  \emph{International conference on machine learning}.\hskip 1em plus 0.5em
  minus 0.4em\relax PMLR, 2018, pp. 1861--1870.

\bibitem{mazzaglia2021contrastive}
P.~Mazzaglia, T.~Verbelen, and B.~Dhoedt, ``Contrastive active inference,''
  \emph{Advances in Neural Information Processing Systems}, vol.~34, 2021.

\bibitem{noel2021online}
A.~D. Noel, C.~van Hoof, and B.~Millidge, ``Online reinforcement learning with
  sparse rewards through an active inference capsule,'' \emph{arXiv preprint
  arXiv:2106.02390}, 2021.

\bibitem{gallagher2000philosophical}
S.~Gallagher, ``Philosophical conceptions of the self: implications for
  cognitive science,'' \emph{Trends in cognitive sciences}, vol.~4, no.~1, pp.
  14--21, 2000.

\bibitem{synofzik2008beyond}
M.~Synofzik, G.~Vosgerau, and A.~Newen, ``Beyond the comparator model: a
  multifactorial two-step account of agency,'' \emph{Consciousness and
  cognition}, vol.~17, no.~1, pp. 219--239, 2008.

\bibitem{friston2012prediction}
K.~Friston, ``Prediction, perception and agency,'' \emph{International Journal
  of Psychophysiology}, vol.~83, no.~2, pp. 248--252, 2012.

\bibitem{lanillos2020robot}
P.~Lanillos, J.~Pages, and G.~Cheng, ``Robot self/other distinction: active
  inference meets neural networks learning in a mirror,'' in \emph{European
  Conference on Artificial Intelligence (ECAI 2020)}, 2020.

\bibitem{nagai2019predictive}
Y.~Nagai, ``Predictive learning: its key role in early cognitive development,''
  \emph{Philosophical Transactions of the Royal Society B}, vol. 374, no. 1771,
  p. 20180030, 2019.

\bibitem{friston2011optimal}
K.~Friston, ``What is optimal about motor control?'' \emph{Neuron}, vol.~72,
  no.~3, pp. 488--498, 2011.

\bibitem{baltieri2020kalman}
M.~Baltieri and C.~L. Buckley, ``On kalman-bucy filters, linear quadratic
  control and active inference,'' \emph{arXiv preprint arXiv:2005.06269}, 2020.

\bibitem{rubinstein1999cross}
R.~Rubinstein, ``The cross-entropy method for combinatorial and continuous
  optimization,'' \emph{Methodology and computing in applied probability},
  vol.~1, no.~2, pp. 127--190, 1999.

\bibitem{theodorou2010generalized}
E.~Theodorou, J.~Buchli, and S.~Schaal, ``A generalized path integral control
  approach to reinforcement learning,'' \emph{The Journal of Machine Learning
  Research}, vol.~11, pp. 3137--3181, 2010.

\bibitem{williams2017model}
G.~Williams, A.~Aldrich, and E.~A. Theodorou, ``Model predictive path integral
  control: From theory to parallel computation,'' \emph{Journal of Guidance,
  Control, and Dynamics}, vol.~40, no.~2, pp. 344--357, 2017.

\bibitem{okada2020variational}
M.~Okada and T.~Taniguchi, ``Variational inference mpc for bayesian model-based
  reinforcement learning,'' in \emph{Conference on Robot Learning}.\hskip 1em
  plus 0.5em minus 0.4em\relax PMLR, 2020, pp. 258--272.

\bibitem{botvinick2012planning}
M.~Botvinick and M.~Toussaint, ``Planning as inference,'' \emph{Trends in
  cognitive sciences}, vol.~16, no.~10, pp. 485--488, 2012.

\bibitem{levine2018reinforcement}
S.~Levine, ``Reinforcement learning and control as probabilistic inference:
  Tutorial and review,'' \emph{arXiv preprint arXiv:1805.00909}, 2018.

\bibitem{millidge2020relationship}
B.~Millidge, A.~Tschantz, A.~K. Seth, and C.~L. Buckley, ``On the relationship
  between active inference and control as inference,'' \emph{arXiv preprint
  arXiv:2006.12964}, 2020.

\bibitem{watson2020active}
J.~Watson, A.~Imohiosen, and J.~Peters, ``Active inference or control as
  inference? a unifying view,'' \emph{arXiv preprint arXiv:2010.00262}, 2020.

\bibitem{watson2020stochastic}
J.~Watson, H.~Abdulsamad, and J.~Peters, ``Stochastic optimal control as
  approximate input inference,'' in \emph{Conference on Robot Learning}.\hskip
  1em plus 0.5em minus 0.4em\relax PMLR, 2020, pp. 697--716.

\bibitem{millidge2021understanding}
B.~Millidge, A.~Tschantz, A.~Seth, and C.~Buckley, ``Understanding the origin
  of information-seeking exploration in probabilistic objectives for control,''
  2021.

\bibitem{deisenroth2011pilco}
M.~Deisenroth and C.~E. Rasmussen, ``Pilco: A model-based and data-efficient
  approach to policy search,'' in \emph{Proceedings of the 28th International
  Conference on machine learning (ICML-11)}.\hskip 1em plus 0.5em minus
  0.4em\relax Citeseer, 2011, pp. 465--472.

\bibitem{lin2000self}
F.~Lin, R.~D. Brandt, and G.~Saikalis, ``Self-tuning of pid controllers by
  adaptive interaction,'' in \emph{Proceedings of the 2000 American Control
  Conference. ACC (IEEE Cat. No. 00CH36334)}, vol.~5.\hskip 1em plus 0.5em
  minus 0.4em\relax IEEE, 2000, pp. 3676--3681.

\bibitem{naiborhu2006direct}
J.~Naiborhu, S.~Nababan, R.~Saragih, and I.~Pranoto, ``Direct gradient descent
  control as a dynamic feedback control for linear system,'' \emph{Bulletin of
  the Malaysian Mathematical Sciences Society}, vol.~29, no.~2, 2006.

\bibitem{sandved2021towards}
L.~Sandved-Smith, C.~Hesp, J.~Mattout, K.~Friston, A.~Lutz, and M.~J. Ramstead,
  ``Towards a computational phenomenology of mental action: modelling
  meta-awareness and attentional control with deep parametric active
  inference,'' \emph{Neuroscience of consciousness}, vol. 2021, no.~1, p.
  niab018, 2021.

\bibitem{Cleeremans-TCS2019}
A.~Cleeremans, D.~Achoui, A.~Beauny, L.~Keuninckx, J.-R. Martin,
  S.~Mu{\~n}oz-Moldes, L.~Vuillaume, and A.~De~Heering, ``Learning to be
  conscious,'' \emph{Trends in cognitive sciences}, vol.~24, no.~2, pp.
  112--123, 2020.

\bibitem{hesp2021deeply}
C.~Hesp, R.~Smith, T.~Parr, M.~Allen, K.~J. Friston, and M.~J. Ramstead,
  ``Deeply felt affect: The emergence of valence in deep active inference,''
  \emph{Neural computation}, vol.~33, no.~2, pp. 398--446, 2021.

\bibitem{tani1998interpretation}
J.~Tani, ``An interpretation of the ‘self’ from the dynamical systems
  perspective: A constructivist approach,'' \emph{Journal of Consciousness
  Studies}, vol.~5, no. 5-6, pp. 516--542, 1998.

\bibitem{tani2020cognitive}
J.~Tani and J.~White, ``Cognitive neurorobotics and self in the shared world, a
  focused review of ongoing research,'' \emph{Adaptive Behavior}, p.
  1059712320962158, 2020.

\bibitem{friston2003dynamic}
K.~J. Friston, L.~Harrison, and W.~Penny, ``Dynamic causal modelling,''
  \emph{Neuroimage}, vol.~19, no.~4, pp. 1273--1302, 2003.

\bibitem{toussaint2009robot}
M.~Toussaint, ``Robot trajectory optimization using approximate inference,'' in
  \emph{Proceedings of the 26th annual international conference on machine
  learning}, 2009, pp. 1049--1056.

\bibitem{deisenroth2013survey}
M.~P. Deisenroth, G.~Neumann, J.~Peters \emph{et~al.}, ``A survey on policy
  search for robotics,'' \emph{Foundations and trends in Robotics}, vol.~2, no.
  1-2, pp. 388--403, 2013.

\end{thebibliography}

\appendix[]
\subsection{Variational Free Energy}
\label{appendix:vfe}
We use the definition of the $\mathcal{F}$ based on \cite{friston2010action}, where the action is implicit within the observation model $\obs(\action)$. Using the KL-divergence the VFE is defined as follows:
\small
\begin{align}
\mathcal{F} &= \text{KL}\left[ q(\latent) || p(\latent|\obs) \right] - \log p(\obs) \nonumber\\
&= \text{KL}\left[ q(\latent) || p(\latent,\obs) \right] = -\text{ELBO} \nonumber
\end{align}
\normalsize
\noindent\textit{Proof.} $\mathcal{F}$ is the KL-divergence between the variational density and the true posterior minus the sensory surprise.
\small
\begin{align}
&\text{KL}\left[ q(\latent) || p(\latent|\obs) \right] = \int_\latent q(\latent) \log \frac{q(\latent)}{p(\latent|\obs)} =\nonumber\\
&=\int_\latent q(\latent) \log q(\latent) \!-\! \int_\latent q(\latent) \log p(\latent|\obs) = \nonumber\\
&= \int_\latent q(\latent) \log q(\latent) \!-\! \int_\latent q(\latent) \log p(\latent,\obs) \!+\! \int_\latent q(\latent) \log p(\obs)= \nonumber\\
&= \int_\latent q(\latent) \log q(\latent) \!-\! \int_\latent q(\latent) \log p(\latent,\obs) \!+\!  \log p(\obs) \int_\latent q(\latent) = \nonumber\\
&= \int_\latent q(\latent) \log q(\latent) \!-\! \int_\latent q(\latent) \log p(\latent,\obs) \!+\!  \log p(\obs) = \nonumber\\
&= \mathcal{F} + \log p(\obs).\nonumber
\end{align}
\normalsize
\noindent\textit{Proof.} The $\mathcal{F}$ is the negative Evidence Lower Bound:
\small
\begin{align}
\mathcal{F} &= \text{KL}\left[ q(\latent) || p(\latent,\obs) \right] \nonumber \\
&= \int_\latent q(\latent) \log q(\latent) \!-\! \int_\latent q(\latent) \log p(\latent,\obs)  \nonumber\\
&=- \left[-\int_\latent q(\latent) \log q(\latent) \!+\! \int_\latent q(\latent) \log p(\latent,\obs)\right] \nonumber\\
&=-\int_\latent q(\latent) \log \frac{p(\latent,\obs)}{q(\latent)} =  - \text{ELBO} \nonumber
\end{align}
\normalsize
\subsection{Laplace Approximation}
\label{appendix:laplace}
\begin{figure}[!hbtp]
    \centering
    \includegraphics[width=0.35\textwidth, height=110px]{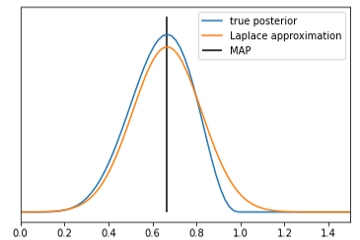}
    \caption{Comparison between the true posterior, and the Laplace approximation and the Maximum A Posteriory (MAP) estimation.}
    \label{fig:obscheme}
\end{figure}

Maximum a posteriori (MAP) approximates the posterior $p(x | y)$ with a point mass (delta function) by simply capturing its mode. MAP is attractive because is fast and efficient. How can we use MAP to construct a better approximation to the posterior? The Laplace approximation is one way of improving a MAP estimate (Fig. \ref{fig:obscheme}) by encoding the posterior with a normal distribution centered at the MAP estimate. 

\vspace{-10px}
\subsection{Mean field Approximation}
\label{appendix:mean-field}
The mean field approximation is a simplifying assumption about the for of the variational distribution $q(\cdot)$. Consider $N$ random variables $x = [x_1, ..., x_N]$ and data $y$. Our aim is to approximate the distribution $p(x | y) $ with a variational distribution $q(x)$. Under the mean-field assumption this could correspond to the following factorization: \small$ p(x | y) \approx q(x) = q(x_1, ..., x_N) =  \prod_{i=1}^N q_i(x_i).$
\normalsize
Essentially, we assume a variational function which partitions the variables into independent parts. This greatly simplifies computations. 


\subsection{Generalized coordinates}
\label{appendix:gen-coords}
A key ingredient behind the success of FEP when compared to other estimators is the use of generalized coordinates, which provides a mathematical framework to model the colored noise in data. The time dependent quantities in the generative model are modeled in generalized coordinates using its higher order derivatives as $\latent = [x \ x' \ x'' \ ...]^T$. Quantitatively, it tracks the probability density of the trajectory of states instead of the point estimates, when compared to the Kalman Filter. To embed the higher order information of the colored noise, the process and observation noises are also modelled in generalized coordinates as $\Tilde{\munoise}$ and $\Tilde{\ynoise}$. Under a Gaussian convoluted white noise assumption, the precision matrix for noise smoothness that represents the cross correlation between each noise derivatives can be written as \cite{friston2008variational}:
\small
\begin{equation}
\label{eqn:Smatrix}
    S(\sigma^2) = \begin{bmatrix}
    1 & 0 & -\frac{1}{2\sigma^2} & .. \\
    0 & \frac{1}{2\sigma^2} & 0 & ..\\
    -\frac{1}{2\sigma^2} & 0 & \frac{3}{4\sigma^4}& ..\\
    .. & .. & .. & ..
    \end{bmatrix}^{-1}_{(p+1)\times (p+1)} \nonumber
\end{equation}
\normalsize
where $\sigma$ is the kernel width of the Gaussian filter and $p$ is the order of generalized motion of states. The generalized noise precision matrices for process noise and observation noise becomes $\Tilde{\Pi}_\munoise = S(\sigma_\munoise^2) \otimes \Pi_\munoise$ and $\Tilde{\Pi}_\ynoise = S(\sigma_\ynoise^2) \otimes \Pi_\ynoise$. The use of generalized coordinates has demonstrated a superior performance in state estimation when compared to the Kalman Filter, and in input estimation when compared to unknown input observers, for colored noise \cite{meera2020free} . The use of generalized coordinates has also proven to improve the state estimation accuracy of a quadrotor hovering in wind \cite{bos2021free}.

\subsection{Expected free energy (EFE)}
\label{appendix:EFE}
The analytical expression for the optimal plan, at the minimum of the EFE over trajectories, can be expressed as a softmax over the path integral of the EFE of each trajectory. Namely, given the EFE over a plan, we have,
\small
    \begin{align}
      \mathcal{G}(\pi) &=  \mathbb{E}_{q(x_{1:T}, y_{1:T}, \pi)}[\ln q(x_{1:T}, \pi) - \ln \tilde{p}(y_{1:T}, x_{1:T}, \pi)] \nonumber\\
        &= D_{KL}[q(\pi)||p(\pi) \sum_\tau^T - \mathbb{E}_{q(o_\tau, x_\tau | \pi)}[\ln q(x_\tau | \pi) - \ln \tilde{p}(o_\tau, x_\tau | \pi)]] \nonumber\\
        &= D_{KL}[q(\pi)||p(\pi) \sum_\tau^T -\mathcal{G}_{\pi_\tau}] \nonumber
\end{align}
\normalsize
Thus, the optimal posterior is given by \small $\underset{\pi}{\mathrm{argmin}} \, \mathcal{G}(\pi)
\implies q^*(\pi) = \sigma(\ln p(\pi) - \sum_\tau^T \mathcal{G}_{\pi_\tau})$
\normalsize
Moreover, the EFE for a specific timestep and plan can be decomposed into the extrinsic and intrinsic value.
\small
\begin{align*}
    \mathcal{G}_\pi &= \mathbb{E}_{q(y_t, x_t)}[\ln q(x_t | \pi) - \ln \tilde{p}(y_t, x_t | \pi)] \\
    &=  \mathbb{E}_{q(y_t, x_t | \pi)}[\ln q(x_t | \pi) - \ln \tilde{p}(y_t, x_t | \pi) \\
    & \ \ \ \ \ \ \ \ \ \ \ \ \ + \ln q(x_t | y_t) - \ln q(x_t | y_t)]\\
    &=-\mathbb{E}_{q(y_t, x_t | \pi)}[\ln \tilde{p}(y_t)] - \mathbb{E}_{q(y_t | \pi)}D_\mathrm{KL}[q(x_t | y_t) || q(x_t | \pi)] 
   \\ &\ \ \ + \mathbb{E}_{q(y_t)}D_\mathrm{KL}[q(x_t | y_t)||p(x_t | y_t \pi)] \\
    &\approx \underbrace{-\mathbb{E}_{q(y_t, x_t | \pi)}[\ln \tilde{p}(y_t)]}_{\text{Extrinsic Value}} - \underbrace{\mathbb{E}_{q(y_t | \pi)}D_\mathrm{KL}[q(x_t | y_t) || q(x_t | \pi)]}_{\text{Intrinsic Value}}
\end{align*}
\normalsize

%








\end{document}